%% file: main_arxiv.tex
\documentclass[a4paper,11pt]{article}
\usepackage[top=2.5cm, bottom=2.5cm, left=2.5cm, right=2.5cm]{geometry}
\usepackage{settings}
\usepackage{dsfont}
\allowdisplaybreaks

\title{
Topological Exploration of High-Dimensional Empirical Risk 
Landscapes: general approach, and applications to phase retrieval}

\author[1]{Antoine Maillard\thanks{antoine.maillard@inria.fr}}
\author[2,3]{Tony Bonnaire}
\author[3]{Giulio Biroli}

\affil[1]{\small  INRIA Paris \& DI ENS, PSL University, Paris, France.}
\affil[2]{\small Université Paris-Saclay, CNRS, Institut d'Astrophysique Spatiale, 91405 Orsay, France}
\affil[3]{\small Laboratoire de Physique de l’École normale supérieure, ENS, Université PSL, CNRS, Sorbonne Université, Université Paris Cité, F-75005 Paris, France}

\begin{document}

\maketitle
\begin{abstract}%
We consider the landscape of empirical risk minimization for high-dimensional Gaussian single-index models (generalized linear models). The objective is to recover an unknown signal $\btheta^\star \in \bbR^d$ (where $d \gg 1$) from a loss function $\hR(\btheta)$ that depends on pairs of labels $(\bx_i \cdot \btheta, \bx_i \cdot \btheta^\star)_{i=1}^n$, with $\bx_i \iid \mcN(0, \Id_d)$, in the proportional asymptotic regime $n \asymp d$. Using the Kac-Rice formula, we analyze different complexities of the landscape---defined as the expected number of critical points---corresponding to various types of critical points, including local minima. We first show that some variational formulas previously established in the literature for these complexities can be drastically simplified, reducing to explicit variational problems over a finite number of scalar parameters that we can efficiently solve numerically. Our framework also provides detailed predictions for properties of the critical points, including the spectral properties of the Hessian and the joint distribution of labels. We apply our analysis to the real phase retrieval problem for which we derive complete topological phase diagrams of the loss landscape, characterizing notably BBP-type transitions where the Hessian at local minima (as predicted by the Kac-Rice formula) becomes unstable in the direction of the signal. We test the predictive power of our analysis to characterize gradient flow dynamics, finding excellent agreement with finite-size simulations of local optimization algorithms, and capturing fine-grained details such as the empirical distribution of labels. 
Overall, our results open new avenues for the asymptotic study of loss landscapes and topological trivialization phenomena in high-dimensional statistical models.

\end{abstract}

\setcounter{tocdepth}{2}
\tableofcontents

\section{Introduction}\label{sec:intro}
\input{sections/introduction.tex}


\section{Landscape analysis via the Kac-Rice approach}\label{sec:kac_rice}
\input{sections/kac_rice_main.tex}

\section{Probing the theory: gradient descent dynamics}\label{sec:simulation_dynamics}
\input{sections/simulations_dynamics.tex}

\newpage
\section{Conclusion}\label{sec:conclusion}
\input{sections/conclusion.tex}

\printbibliography

\newpage
\appendix 
\addtocontents{toc}{\protect\setcounter{tocdepth}{1}} 

\section{The Kac-Rice formula: technicalities and numerical solutions}\label{sec_app:theory_kac_rice}
\input{sections/appendix/theory_kac_rice.tex}

\section{Derivation of the BBP transition condition}\label{sec_app:bbp}
\input{sections/appendix/bbp.tex}

\section{Further exploration of the phase retrieval landscape}\label{sec_app:more_kac_rice}
\input{sections/appendix/more_kac_rice.tex}

\section{Extension of the dynamical comparison to $a=1.0$}\label{sec_app:dynamics_general_a}
\input{sections/appendix/dynamics_general_a.tex}

\end{document}

%% file: sections/introduction.tex
Minimization of non-convex, high-dimensional landscapes with potentially many local minima is routinely carried out in machine learning and optimization with remarkable success using simple local iterative procedures such as gradient descent and its stochastic variants. Understanding why, and under what conditions, these algorithms work so well remains a major open challenge, and has motivated a large body of recent work. Without aiming to be exhaustive, representative contributions include \cite{Neyshabur2017, Belkin2018, ma2018power, Venturi2019, Mannelli2020a, Martin2023, Annesi2024}. 

Over the past decade, considerable effort has been devoted to analyzing the geometry of loss landscapes—focusing on properties such as the presence of spurious minima, saddle points, and connectivity—and to elucidating their relationship with training dynamics. Several works have shown that, in specific parameter regimes, spurious local minima disappear, for instance when the signal-to-noise ratio is sufficiently large \cite{Soudry2016, Cai2022}. In such cases, despite their intrinsic non-convexity, landscapes effectively become easy to optimize. This observation has led to explanations of the success of gradient-based methods in terms of a “trivialization” of the loss landscape \cite{Fyodorov2004}, associated with the absence of bad minima. Yet this perspective leaves out a salient empirical fact: substantial evidence now shows that suboptimal minima may still exist even in parameter regimes where optimization performs well in practice \cite{Baity-Jesi2019, liu2020bad}.

Among the various approaches developed to study high-dimensional loss landscapes, the one most closely related to the present work is the rigorous characterization of their geometric structure. This line of research \cite{Fyodorov2004,Auffinger2010} originated in the study of spin-glass models—physical systems characterized by highly rugged energy landscapes—and was later extended to problems in statistical inference and machine learning~\cite{ge2017optimization,maillard2020landscape,asgari2025local}. The main analytical tool in this context is the Kac–Rice method, which has been reviewed recently in \cite{ros}. Using this framework, Gaussian high-dimensional landscapes have been thoroughly characterized, including those arising in spherical spin-glass models and inference problems such as tensor PCA \cite{Ros2019,Arous2019}. Connections between landscape geometry and training dynamics have also been investigated using statistical-physics techniques \cite{Mannelli2019a,Mannelli2019,SaraoMannelli2020}.

In contrast, typical loss landscapes encountered in machine learning are defined as empirical sums over training data and are therefore not Gaussian. Extensions of the Kac–Rice method to this more general setting were first developed in \cite{maillard2020landscape} for generalized linear models, and further advanced in the recent work \cite{asgari2025local} to multi-index models. Nevertheless, a detailed characterization of the geometry of non-convex landscapes arising in generalized linear models, and of its evolution as a function of the signal-to-noise ratio, is still lacking. Given the large number of recent studies devoted to the training dynamics of generalized linear models, see for instance \cite{collins2024hitting,bietti2023learning,ben2022high,lee2024neural,arnaboldi2023escaping,montanari2026phase}, this represents a timely and important challenge. Addressing it would make it possible to establish a theoretical connection between loss-landscape geometry and optimization dynamics beyond the Gaussian case, and in settings directly relevant to realistic machine learning models.

The aim of this work is to initiate a systematic investigation along this direction. We focus on phase retrieval, a generalized linear model of practical relevance that has been extensively studied from the perspective of training dynamics. Our approach can be straightforwardly generalized to other generalized linear models.

\subsection{The setting: single/multi-index models and phase retrieval}

In this work we focus on learning a single-index model through empirical risk minimization. 
Formally, we consider a general form of empirical risk function:
\begin{align}\label{eq:loss}
    \hR(\btheta) \coloneqq \frac{1}{n} \sum_{i=1}^{n} \ell(\bx_i \cdot \btheta, \bx_i \cdot \btheta^\star).
\end{align}
The loss function 
$\ell : \bbR^2 \to \bbR$ is taken so far to be smooth but generic.
The parameters $\btheta \in \bbR^d$ and the ground-truth $\btheta^\star \in \bbR^d$ are taken to be unit normed
$\|\btheta\|_2 = \|\btheta^\star\|_2 = 1$, and we assume that the \emph{data/sensing vectors} $\{\bx_i\}_{i=1}^n$ are taken i.i.d.\ from the standard Gaussian distribution $\mcN(0, \Id_d)$. 
We consider the minimization of $\hR(\btheta)$ in a high-dimensional regime where both the number of parameters $d$ and the sample size $n = n(d)$ go to infinity, at a fixed ratio:
\begin{align}\label{eq:def_alpha}
    \lim_{d \to \infty} \frac{n}{d} \to \alpha > 1.
\end{align}

\myskip
While we will state our theoretical results for a generic loss function $\ell$, 
the main focus of our applications is the so-called \emph{phase retrieval}\footnote{Which is a slight abuse: we rather are considering a ``sign retrieval'' problem since variables are real.} problem,  which is relevant for accurately reconstructing signals and images and enabling reliable measurements in fields like optics, imaging, and quantum science~\cite{dong2023phase}.
Concretely, following \cite{bonnaire2025role,cai2021globalI,cai2021globalII}, we consider the following loss function 
\begin{align}\label{eq:def_ell_a}
    \ell_a(y, y^\star) \coloneqq \frac{(y^2 - [y^\star]^2)^2}{a + [y^\star]^2},
\end{align}
for some normalization parameter $a > 0$. The reason behind this choice is to evaluate a relative error, whereas without the denominator rare but very large $[y^\star]^2$ can have a strong influence on the loss. Further, this loss function has second derivative (w.r.t.\ $y$) bounded from below, which will be a useful property in what follows.

\subsection{Summary of contributions}

The main contributions of this work are as follows.

\myskip
\textbf{A systematic framework for loss landscape analysis beyond Gaussian settings --}
We initiate a systematic investigation of the geometry of non-convex empirical risk landscapes in high dimension, going substantially beyond what was previously available.
Building on the Kac--Rice approach of~\cite{maillard2020landscape,asgari2025local}, we derive \emph{tractable scalar variational formulas} for the so-called annealed complexities\footnote{i.e.\ the average number of these points.} of generic critical points, saddles of sub-extensive index, and an upper bound for the one of local minima.
While we focus on phase retrieval as our illustrating example, we emphasize that our approach is \emph{not} limited to this problem: it applies broadly to other single-index models (e.g.\ logistic regression, other generalized linear models), and can be extended to multi-index models using the framework of~\cite{asgari2025local}, or to other high-dimensional estimation tasks such as Gaussian mixture classification, as discussed in~\cite{maillard2020landscape}.
Furthermore, although in this work we restrict to parameters constrained to the unit sphere $\mcS^{d-1} \coloneqq \{\btheta \in \bbR^d \, : \, \|\btheta\| = 1\}$, the same methodology can be applied to other geometries and regularization schemes, as considered in~\cite{asgari2025local}.

\myskip
\textbf{Detailed characterization of critical points --}
A key feature of our approach is that the variational formulas yield not only the complexities of the different types of critical points, but also precise predictions for their statistical properties, more precisely we obtain:
\begin{itemize}
    \item The \emph{spectral density of the Hessian} of $\hR(\btheta)$ at a typical critical point of a given type (local minimum, sub-extensive-index saddle, or generic saddle), possibly at a fixed value of the loss (or ``energy'') $\hR(\btheta)$, and fixed overlap $q \coloneqq \btheta \cdot \btheta^\star$ with the signal;
    \item The limiting \emph{empirical distribution of the labels} $(y_i, y_i^\star)_{i=1}^n = (\bx_i \cdot \btheta, \bx_i \cdot \btheta^\star)_{i=1}^n$, at typical critical points of a given type (as above).
\end{itemize}
Our characterization of local minima 
goes through a complexity measure $\tSigma_0$, which is a tighter upper bound on the annealed complexity of local minima than the sub-extensive-index complexity (as studied in~\cite{asgari2025local}). We add to it an additional constraint, encoding the requirement that the signal direction is not a descent direction at a local minimum: it proves crucial, as it enables us to establish a finite trivialization threshold $\alpha_\triv$ for the landscape of phase retrieval, while we find that the complexity $\Sigma_\fin$ of sub-extensive-index saddles never trivializes for any finite $\alpha$.

\myskip
\textbf{A BBP--Kac-Rice method connecting landscape geometry to dynamics --}
We combine the Kac--Rice analysis with random matrix theory (specifically, the Baik--Ben Arous--P\'ech\'e transition~\cite{baik2005phase,benaych2011eigenvalues}) to determine whether a negative outlier eigenvalue--aligned with the signal--emerges from the Hessian at a typical critical point. 
This ``BBP--Kac-Rice'' method extends the analysis of~\cite{bonnaire2025role} from the equator ($q = 0$) to arbitrary overlaps $q$. It reveals that the instability of local minima towards the signal can occur in a wide range of $\alpha$ where the Kac--Rice upper bound still predicts a positive complexity, providing an accurate prediction of the algorithmic success of gradient descent. This allows us to compute precise topological phase diagrams for phase retrieval in the $(\alpha, q)$ plane, identifying the thresholds for trivialization of local minima, sub-extensive-index saddles, and generic critical points, as well as BBP instability thresholds for high-energy, typical-energy, and low-energy minima.

\myskip
\textbf{Numerical analysis of gradient descent dynamics and comparison with analytical landscape predictions --}
We present a thorough numerical study of gradient descent using full-batch gradient-descent (GD) simulations at dimension $d = 512$ and for many $(\alpha, q)$, and compare the results to the analytical landscape predictions.
A key challenge in such comparisons is the presence of strong finite-size corrections to the infinite-dimensional limit, as shown in~\cite{bonnaire2025role}. To circumvent these corrections while faithfully probing the large-$d$ limit, we adopt a modified dynamics introduced in~\cite{bonnaire2025role} and extend it to arbitrary overlaps $q > 0$: the dynamics is first run under a constraint that pins the overlap $\btheta \cdot \btheta^\star$ to a prescribed value $q_0$ for a burn-in phase, before being released to evolve freely.

\myskip
The numerical comparison yields the following results:
\begin{itemize}
    \item \textbf{Energy bands --} For $q \in \{0.0, 0.1, 0.2\}$ and a range of $\alpha$, the average energies of minima in which GD gets trapped lie consistently inside the energy band predicted by the Kac--Rice method, and closely track the \emph{typical} energy $e_\star = \argmax_e \tSigma_0(q, e)$. 
    \item \textbf{Hessian spectral density --} The empirical distribution of Hessian eigenvalues at  minima in which GD gets trapped is in excellent agreement with the annealed Kac--Rice prediction $\rho_0$, both in the bulk and in the tails, for all tested values of $(\alpha, q)$. 
    \item \textbf{Distribution of second derivatives and joint label distribution --} The empirical distribution of the second derivatives $F(y_i, y_i^\star) \coloneqq \partial_1^2 \ell(y_i, y_i^\star)$ at  minima in which GD gets trapped matches the theoretical prediction closely. The joint law $\nu(y, y^\star)$ of predicted and ground-truth labels is also visually indistinguishable from the annealed Kac--Rice prediction $\nu_0$. Crucially, both theory and experiment exhibit a markedly non-Gaussian structure: even at $q = 0$ (zero overlap with the signal), the predicted labels $y_i = \btheta \cdot \bx_i$ align closely with the ground-truth labels $y_i^\star = \btheta^\star \cdot \bx_i$, in sharp contrast to what a generic random configuration would produce. 
    \item \textbf{Algorithmic phase diagram --} We overlay the empirical GD success rates in the $(\alpha, q)$ plane on top of the theoretical phase diagram. For uncorrelated initialization, the empirical transition is in very good agreement with the BBP--KR instability threshold for typical-energy minima. Extending this comparison to the full $(\alpha, q)$ plane, the empirical success boundary aligns with the theoretical prediction but also shows some discrepancy which we link to numerical difficulties in establishing the correct algorithmic threshold at increasingly larger values of $q$, and possibly to the need for a quenched Kac-Rice computation \cite{Ros2019}. 
\end{itemize}
Taken together, these comparisons demonstrate that the theoretical framework we develop provides a highly accurate and interpretable picture of the minima encountered by gradient descent in large dimensions. We plan to release an extended version of this work, which will include a complementary analysis based on replica-symmetry breaking (1RSB) theory, as well as further numerical experiments exploring the Kac--Rice predictions across a broader range of models and parameters.


\myskip 
The analytical results presented in this work are derived at the level of rigor customary in theoretical physics. We believe that a fully rigorous mathematical proof of our results is within reach by leveraging the analysis of~\cite{maillard2020landscape,asgari2025local}, but we leave this task for future research. We have intentionally adopted a presentation tailored to mathematically inclined readers, with the aim of stimulating interest within the mathematics community and encouraging the development of this line of research on a fully rigorous basis.
More specifically, we begin from the annealed formulas for the complexities, which are essentially rigorous thanks to~\cite{maillard2020landscape,asgari2025local}. The simplification into scalar variational principles is done at the theoretical physics level of rigor. While we expect that these results can be established rigorously, we defer this question to future work. 
Finally, while we can not guarantee mathematically that these scalar variational principles do not admit several solutions, we numerically never found more than one solution for local minima and saddles of sub-extensive index. This is however not the case for the complexity of all critical points, for which we exhibited the presence of two fixed points in a (very limited) region of parameters, associated to a first-order phase transition, see Appendix~\ref{subsec_app:phase_transition_total_complexity}.

\subsection{Related works}


There are only a few works applying the Kac–Rice method to non-Gaussian landscapes, such as those arising from empirical risk functions. The first analysis of this type was carried out in \cite{maillard2020landscape}, which the present work builds upon. Shortly thereafter, \cite{fyodorov2022optimization} extended the Kac–Rice framework beyond purely Gaussian random landscapes. More recently, \cite{tsironis2025landscape} derived equations for the average number of critical points in perceptron and generalized linear models in the presence of structured data, and \cite{cai2021globalI,cai2021globalII} showed  that trivialization occurs in phase retrieval with the loss of eq.~\eqref{eq:def_ell_a} at large enough $\alpha$ (not scaling with $n$).
Even more relevant for our work, \cite{asgari2025local} applies the Kac-Rice method to high-dimensional estimation. In particular, it generalizes the results of~\cite{maillard2020landscape} to so-called multi-index models (and removes technical assumptions used in~\cite{maillard2020landscape}) and derives sharp results in convex settings 
using the Kac-Rice formalism.
\myskip 

The instability of local minima induced by a BBP transition toward the signal, and its implications for gradient-flow dynamics, were first discussed in \cite{SaraoMannelli2020,Mannelli2019} in the context of tensor–matrix PCA, and later extended to phase retrieval in \cite{Mannelli2020b}. Our work is closely related to the recent study \cite{bonnaire2025role}, which investigates the role of the BBP transition in the time-dependent Hessian for phase retrieval. See also the very recent preprints \cite{montanari2026phase}, which analyze a broader class of models, and \cite{annesi2025overparametrization}, which highlights the role of over-parameterization in the BBP transition at the beginning of the dynamics.


\subsection{Organization of the paper}

Section~\ref{sec:kac_rice} is dedicated to the presentation of the Kac-Rice approach:
we introduce the principles of the method and its application to single-index models, as well as tractable asymptotic formulas for the complexities of different types of saddle points in the high-dimensional limit. 
We finish with a thorough numerical exploration of these predictions, showing in particular predictions for the spectral density of the Hessian, for the joint law of the labels $(\bx_i \cdot \btheta, \bx_i \cdot \btheta^\star)$, and trivialization transitions in the landscape.

\myskip
In Section~\ref{sec:simulation_dynamics}, we probe our theory by comparing the Kac-Rice predictions to the results of finite-size simulations of gradient descent dynamics. 
We show a strong qualitative agreement in terms of thresholds, and a very good agreement in terms of predictions of the laws of the labels and of the spectrum of the Hessian.

\myskip
We finally discuss the outcomes of our analysis in Section~\ref{sec:conclusion}, as well as its limitations and future directions. 

\subsection{Notations}

Throughout this paper, we use the following notations: 
\begin{itemize}[leftmargin=*]
    \item $\mcS^{d-1}$: Euclidean unit sphere in $\bbR^d$.
    \item $\grad, \Hess$: Riemannian gradient and Hessians on the sphere $\mcS^{d-1}$.
    \item Index ${\rm i}(H)$: number of negative eigenvalues of a matrix $H$.
    \item $\partial_1 \ell, \partial_1^2 \ell$: successive derivatives of $\ell$ with respect to its first variable.
    \item $\mcP(E)$: probability measures on $E$.
\end{itemize}

%% file: sections/kac_rice_main.tex
We describe here our main theoretical results regarding the landscape complexity of generalized linear models.
After introducing the setting and our main goals in Section~\ref{subsec:setting}, we sketch in Section~\ref{subsec:kac_rice_sketch} the analytical derivation of the asymptotics of the average log-number of critical points (so-called \emph{complexities}) using the Kac-Rice method. This approach was developed in~\cite{maillard2020landscape,asgari2025local} in our context: it yields variational formulas over spaces of probability measures for these asymptotic complexities, and provides bounds on the average log-number of local minima.
These formulas are presented in Section~\ref{subsec:main_th_kr}, where we also show how they can be greatly reduced to variational formulas over a finite number of scalar parameters: we then present simple algorithms for approximating their solution.
The Kac–Rice method provides access to the bulk eigenvalue density of the Hessian at critical points, but it does not capture potential outlier eigenvalues associated with directions aligned with the signal. To investigate whether such outliers emerge as a function of the signal-to-noise ratio, we apply and extend the analysis of \cite{bonnaire2025role}. This approach allows us to determine whether a Baik–Ben Arous–Péché (BBP) transition occurs as a function of $\alpha$, for typical critical points and minima at fixed loss value and fixed overlap with the signal.
Finally, in Section~\ref{subsec:exploring_landscape} we leverage these algorithms to numerically explore the landscape complexities in high-dimensional (real) phase retrieval, a flagship example of a non-convex landscape in high-dimensional statistics.  We probe the properties of various types of critical points, such as the Hessian's spectral density, and present sharp results for the trivialization of the landscape (i.e.\ the disappearance of all local minima) as a function of the sample complexity.

\subsection{Setting and objectives of the Kac-Rice method}\label{subsec:setting}

\subsubsection{Critical points and complexities}\label{subsubsec:complexities}

For any $Q \subseteq (-1,1)$ and $B \subseteq \bbR$ two non-empty open intervals, let us denote 
\begin{align}
    \mcE(Q, B) \coloneqq \{\btheta \in \mcS^{d-1} \, : \, \btheta \cdot \btheta^\star \in Q \, \textrm{ and } \hR(\btheta) \in B\}
\end{align}
We will separate the counting of different types of critical points in the landscape, according to the 
number of descending directions in the Hessian:
\begin{align}\label{eq:def_Sigma_tot_fin}
    \begin{dcases}
    \Sigma_{\rm tot.}(Q, B) &\coloneqq \lim_{d \to \infty} \frac{1}{d} \log \EE \#\{\btheta \in \mcE(Q, B) \, : \, \grad \, \hR(\btheta) = 0 \}, \\
    \Sigma_{\rm k}(Q, B) &\coloneqq \lim_{d \to \infty} \frac{1}{d} \log \EE \#\{\btheta \in \mcE(Q, B) \, : \, \grad \, \hR(\btheta) = 0 \, \textrm{ and } \mathrm{i}[\Hess \, \hR(\btheta)] \leq k\}.
    \end{dcases}
\end{align}
These complexities count respectively the critical points
and the saddles of index at most $k$ as $d \to \infty$ (i.e.\ with at most $k$ descending directions), that have \emph{overlap} $\btheta \cdot \btheta^\star \in Q$ with the signal, and \emph{loss, or energy}\footnote{We use loss or energy interchangeably to denote the value of $\hR$.}, value $\hR(\btheta) \in B$.
Importantly, $\Sigma_0(Q, B)$ is the complexity of \emph{local minima}, of particular relevance to the training dynamics.
Notice that 
\begin{align*}
    \Sigma_\sub(Q, B) \coloneqq \lim_{\eps \to 0}\Sigma_{\eps d}(Q, B)
\end{align*}
counts critical points of \emph{sub-extensive index} as $d \to \infty$ (i.e.\ with a $\smallO(d)$ number of descending directions).
With respect to other works, $\Sigma_\tot$ was analyzed first in~\cite{maillard2020landscape}, while $\Sigma_{\fin}$ is the quantity characterized in~\cite{asgari2025local} as a proxy for the complexity of local minima.

\myskip
\textbf{A bound on the complexity of minima --}
Finally, we define a refined bound for the complexity of minima, by counting critical points for which $(i)$ the Hessian has a sub-extensive index, 
and $(ii)$ the signal $\btheta^\star$ is \emph{not a descending direction}, i.e.\ $(\btheta^\star)^\T [\Hess \, \hR(\btheta)] \btheta^\star \geq 0$.
\begin{align}\label{eq:def_tSigma0}
        \tSigma_0(Q, B) \coloneqq \lim_{\eps \to 0} \lim_{d \to \infty} \frac{1}{d} \log \EE \,\#\{&\btheta \in \mcE(Q, B) \, : \, \grad \, \hR(\btheta) = 0 \, \textrm{ and } \mathrm{i}[\Hess \, \hR(\btheta)] \leq \eps d \\ 
        \nonumber
        &\, \textrm{ and } (\btheta^\star)^\T [\Hess \, \hR(\btheta)] \btheta^\star \geq 0\}.
\end{align}
Notice that we have the series of bounds.
\begin{align}\label{eq:series_bounds_complexities}
    \Sigma_0(Q, B) \leq \tSigma_0(Q, B) \leq \Sigma_{\fin.}(Q, B) \leq \Sigma_{\rm tot.}(Q, B).
\end{align}
In what follows, we provide exact asymptotics for $(\tSigma_0, \Sigma_\fin, \Sigma_\tot)$. As we will see, the upper bound $\Sigma_0 \leq \tSigma_0$ 
is already sufficient to establish a topological trivialization of the landscape (i.e.\ $\tSigma_0(Q, B) < 0$ for any $Q, B$) at a \emph{finite value of $\alpha = n/d$}.

\subsection{Sketch of the derivation}\label{subsec:kac_rice_sketch}

Here we sketch the derivation of asymptotic formulas for $(\tSigma_0, \Sigma_\fin, \Sigma_\tot)$.
It uses the Kac-Rice formula (see e.g.~\cite{azais2009level,adler2007random}), 
and is a straightforward adaptation of the computation detailed in~\cite{maillard2020landscape}, and generalized in~\cite{asgari2025local}.
For pedagogical reasons, we present a brief and heuristic sketch of the derivation in the case of $\Sigma_\tot(Q, B)$, hence following closely~\cite{maillard2020landscape}, to which we refer for more details.
We indicate in the end how to generalize this computation to $\Sigma_\fin$ and $\tSigma_0$. We note that the computation of~$\Sigma_\fin$ can be seen as a special case of the results of~\cite{asgari2025local}.


\myskip 
\textbf{The Kac-Rice formula --}
The first step of the computation is to apply the celebrated Kac-Rice formula~\cite{kac1943correction,rice1945mathematical} to compute the 
first moment of the number of critical points satisfying certain properties. 
Denote 
\begin{equation*}
        \mcC_{d}(Q,B) \coloneqq \#\{\btheta \in \mcE(Q, B) \, : \, \grad \, \hR(\btheta) = 0\}.
\end{equation*}
The Kac-Rice formula reads (see~\cite{adler2007random,azais2009level}, or~\cite[Lemma~3]{maillard2020landscape} for a detailed mathematical statement in our precise setting):
\begin{equation}\label{eq:KR_1}
 \EE \, \mcC_{d} (Q,B)= \int_{\mcS^{d-1}} \varphi_{\mathrm{grad} \, \hat R(\btheta)}(0)  \, \EE \left[\indi_{\{\btheta \cdot \btheta^\star \in Q ; \hR(\btheta) \in B\}}\left|\det \mathrm{Hess} \hat R(\btheta)\right| \Big| \mathrm{grad}\, \hat R(\btheta) = 0\right] \sigma(\rd \btheta),
\end{equation}
where $\sigma$ is the usual uniform surface measure on $\mcS^{d-1}$, and $\varphi_{\grad \hR(\btheta)}$ is the density of the gradient (importantly this is the spherical gradient, which lives in the tangent space $T_\btheta\mcS^{d-1} \simeq \bbR^{d-1}$).

\myskip 
\textbf{The law of the Hessian and gradient --}
From eq.~\eqref{eq:loss}
we can write the spherical gradient and Hessian as (recall $y_i = \btheta \cdot \bx_i$ and $y_i^\star = \btheta^\star \cdot \bx_i$): 
\begin{align}\label{eq:grad_Hess}
    \begin{dcases}
        \grad \, \hR(\btheta) &= \frac{1}{n} \sum_{i=1}^n (P_{\btheta}^\perp \bx_i) \, \partial_1 \ell(y_i, y_i^\star), \\
        \Hess \, \hR(\btheta) &= \frac{1}{n} \sum_{i=1}^n (P_{\btheta}^\perp \bx_i) (P_{\btheta}^\perp \bx_i)^\T \, \partial_1^2 \ell(y_i, y_i^\star) -  
        \left(\frac{1}{n} \sum_{i=1}^n y_i \, \partial_1 \ell(y_i, y_i^\star)\right)P_\btheta^\perp.
    \end{dcases}
\end{align}
Here $P_\btheta^\perp$ is the orthogonal projector on $T_\btheta \mcS^{d-1}$.

\myskip
\textbf{Conditioning on the law of the labels --}
Crucially, for a fixed value $q = \btheta \cdot \btheta^\star$, the joint law of $(\grad \hR(\btheta), \Hess \hR(\btheta))$ depends only on the empirical law 
of the labels $(y_i, y^\star_i)_{i=1}^n$. 
%
Notice that for a fixed $q$, these variables are sampled i.i.d.\ from $\mu_q$, 
where 
\begin{align}\label{eq:def_muq}
    \mu_q \coloneqq \mcN\left(0, \begin{pmatrix}
        1 & q \\ 
        q & 1
    \end{pmatrix}\right).
\end{align}
Conditioning on $(\bu_i)_{i=1}^n \coloneqq (y_i, y^\star_i)_{i=1}^n$, one can then obtain from eq.~\eqref{eq:KR_1} (we refer to~\cite{maillard2020landscape} for more details):  
\begin{align}\label{eq:KR_2}
 \EE \, \mcC_{d} (Q,B) = \int_Q \rd q \, \omega_d(q) \, \EE_{\bu_i \iid \mu_q} \left[\varphi_{g(\bU)}(0)  \, \delta[C(\bU)] \, \indi_{\hR(\bU) \in B} 
 \EE_{\{\bz_i\}}\left[\left|\det H(\bU)\right| \middle| g (\bU) = 0\right]\right].
\end{align}
In eq.~\eqref{eq:KR_2} we denoted $\bU \coloneqq (\bu_i)_{i=1}^n$. Moreover: 
\begin{enumerate}[label=$( \roman* )$, leftmargin=*]
    \item $\omega_d(q)$ is a volumetric factor (arising from the volume integral in eq.~\eqref{eq:KR_1}). It can be easily computed as:
    \begin{equation*}
        \omega_d(q) = c_d \exp\left\{\frac{d}{2} \left(1+ \log \alpha + \log(1-q^2)\right)\right\},
    \end{equation*}
    where $(1/d) \log c_d \to 0$ as $d \to \infty$.
    \item We denoted $\hR(\bU) = (1/n) \sum_{i=1}^n \ell(y_i, y^\star_i)$, and 
    \begin{equation*}
        \mcC(\bU) \coloneqq \frac{1}{n} \sum_{i=1}^n \left(\frac{P_\btheta^\perp \btheta^\star}{\|P_\btheta^\perp \btheta^\star \|^2} \cdot P_{\btheta}^\perp \bx_i\right) \partial_1 \ell(y_i, y_i^\star) = \frac{1}{n} \sum_{i=1}^n \frac{(y^\star_i - qy_i)}{\sqrt{1-q^2}} \partial_1 \ell(y_i, y_i^\star).
    \end{equation*} 
    Notice that $\mcC(\bU)$ is the projection of the gradient in the direction of $ P_\btheta^\perp \btheta^\star$. Intuitively, one needs to impose that the gradient in this direction is zero since, as detailed below, $g(\bU)$ is the component of the gradient only in the $\{\btheta, \btheta^\star\}^\perp$ space.
    \item Finally, for $\bz_i \iid \mcN(0, \Id_{d-2})$, we have 
    \begin{equation}\label{eq:def_gH}
        \begin{dcases}
            g(\bU) &\coloneqq \frac{1}{n} \sum_{i=1}^n \partial_1 \ell(y_i, y_i^\star) \bz_i , \\
            H(\bU) &\coloneqq \frac{1}{n} \sum_{i=1}^n 
            \begin{pmatrix}
            b_i^2 & b_i \bz_i^\T \\
            b_i \bz_i & \bz_i \bz_i^\T
            \end{pmatrix}\partial_1^2 \ell(y_i, y^\star_i)
            - \left(\frac{1}{n} \sum_{i=1}^n y_i \, \partial_1 \ell(y_i, y_i^\star)\right) \Id_{d-1} ,
        \end{dcases}
    \end{equation}
        with $b_i \coloneqq (y_i^\star - qy_i) / \sqrt{1-q^2}$.
    Note that $g(\bU)$ is the projection of the gradient in $\{\btheta, \btheta^\star\}^\perp$. In $H(\bU)$ we explicitly separated the direction $\btheta^\star$ (corresponding to the first direction in the block-decomposition of $H$) from $\{\btheta, \btheta^\star\}^\perp$.
\end{enumerate}


\myskip 
\textbf{Concentration of the determinant, and large deviations --}
The key next step to simplify eq. (\ref{eq:KR_2}) is the computation of $\EE_{\bz_i}\left[\left|\det H(\bU)\right| \middle| g (\bU) = 0\right]$. We first notice that the conditioning at fixed $y_i,y^\star_i$ reduces to a linear conditioning on $\bz_i$, which can be easily taken into account and leads to a slightly modified random matrix $H'(\bU)$ which is obtained from $H'(\bU)$ by doing the replacement: 
\[
\bz_i \rightarrow \bz_i-\frac{\partial_1 \ell(y_i, y_i^\star)\sum_j \partial_1 \ell(y_j, y_j^\star) \bz_j}{\sum_j (\partial_1 \ell(y_j, y_j^\star))^2 }
\,\,.\]
The key feature of $H'(\bU)$ is that it is simply a \emph{low-rank perturbation} of $H(\bU)$.
Following \cite{maillard2020landscape}, we assume that the large deviations of the spectral distribution of $H'(\bU)$ has rate $d^2$, which is the usual scaling for this kind of random matrices \cite{arous1997large,guionnet2022rare}. Using this result one obtains: 
\begin{equation}
    \lim_{d\rightarrow \infty}\frac{1}{d}\log \EE_{\bz_i}\left[\left|\det H'(\bU)\right| \right]=\lim_{d\rightarrow \infty}\frac{1}{d} \EE_{\bz_i}\left[\log\left|\det H'(\bU)\right| \right]
\end{equation}
The right hand side can be expressed in terms of the spectral density of $H'(\bU)$, which coincides with the one of $H(\bU)$ (since these two matrices only differ by a low-rank term). In consequence, one obtains the final simple expression: 
\begin{equation}\label{eq:def_kappa}
    \kappa_{\alpha}(\nu) \coloneqq \lim_{d\rightarrow \infty}\frac{1}{d} \EE_{\bz_i}\left[\log\left|\det H'(\bU)\right| \right]=\int \mu_{\alpha}[\nu](\rd w) \log |w - t(\nu)|\,,
\end{equation}
where $t(\nu) \coloneqq \int \nu(\rd y, \rd y^\star) \, y \, \partial_1 \ell(y, y^\star)$ and, moreover, $\mu_{\alpha}[\nu]$ is the asymptotic spectral distribution of $\bz \bD \bz^\T / n$, with $\bz \in \bbR^{d \times n}$ with i.i.d.\ $\mcN(0,1)$ elements, and 
$\bD = \Diag(\{ \partial^2_1 \ell(\bu_i)\}_{i=1}^n)$, for $\{\bu_i\}_{i=1}^n \iid \nu$. Eq.~\eqref{eq:def_kappa} is valid for any $\bU$ whose empirical distribution $(1/n) \sum_{i=1}^n \delta_{\bu_i}$ converges, as $n \to \infty$, to the probability measure $\nu$.

\myskip 
Coming back to eq.~\eqref{eq:KR_2},
since both $\omega_d(q)$ and $\mathbb{E}_{\mathbf z_i}\!\left[ \left| \det H(\mathbf U) \right| \,\middle|\, g(\mathbf U)=0 \right]$ scale exponentially with $d$, eq.~\eqref{eq:KR_2} can be evaluated using Laplace's method on $q$, and Sanov’s large deviation principle and Varadhan's lemma\footnote{For a rigorous proof, one would also need to control the constraint involved by $\delta[C(\bU)]$ at a finer level that what is sketched in~\cite{maillard2020landscape}. At the heuristic level of our presentation, one can replace $\delta[C(\bU)]$ by $(2\eps)^{-1} \indi\{|C(\bU)| \leq \eps\}$ for a small $\eps > 0$, which one takes to $0$ after $d \to \infty$. This issue is resolved in the more general results of~\cite{asgari2025local}: we nevertheless present here the derivation of~\cite{maillard2020landscape}, as it matches more closely our setup.
} on $\nu \coloneqq (1/n) \sum_{i=1}^n \delta_{\bu_i}$~\cite{dembo2009large}. 
Combining it with the different results mentioned above,
this allows to obtain the result for the complexity of critical points which was derived in~\cite{maillard2020landscape}\footnote{\cite{maillard2020landscape} uses $\nu$ to denote the asymptotic law 
of $\left(y_i, [y_i^\star - q y_i] / \sqrt{1-q^2}\right)_{i=1}^n$. Here we use $\nu$ to denote the asymptotic law of $(y_i, y_i^\star)_{i=1}^n$, which induces a slight change in notations.}:
\begin{align}\label{eq:Sigma_TC_a}
    \nonumber
    \Sigma_\TC(Q, B) &= \sup_{q \in Q}\sup_{e \in B} \Bigg\{\frac{1 + \log \alpha}{2} + \frac{1}{2} \log (1-q^2) \\ 
    &\hspace{10pt}+ \sup_{\nu \in \mcM_\ell(q, e)}\left[-\frac{1}{2} \log \int_{\bbR^2} \nu(\rd \bu) A(\bu)
    + \kappa_{\alpha}(\nu) - \alpha H(\nu | \mu_q)\right]\Bigg\}.
\end{align}
where $\mcM_\ell(q, e)$ is the set of probability measures on $\bbR^2$ such that $\int \nu(\rd \bu) \ell(\bu) = e$ and 
$\int \nu(\rd \bu) c_q(\bu) = 0$, with $\bu \coloneqq (y, y^\star)$.
Secondly, $H(\nu | \mu) \coloneqq \int \log(\rd \nu / \rd \mu) \rd \nu$ is the relative entropy, or Kullback-Leibler divergence, which appears from the use of Sanov's large deviation principle.
Finally, recall that $\kappa_{\alpha}(\nu)$ is given in eq.~\eqref{eq:def_kappa}, and we defined the two functions:
\begin{align}\label{eq:def_Ac}
    \begin{dcases}
    A(y,y^\star) &\coloneqq (\partial_1 \ell[y, y^\star])^2, \\
    c_q(y,y^\star) &\coloneqq \frac{y^\star - q y}{\sqrt{1-q^2}}\cdot \partial_1 \ell[y, y^\star].
    \end{dcases}
\end{align}

\myskip 
\textbf{Generalization to $\Sigma_\fin$ and $\tSigma_0$ --}
Let us now detail how to generalize the computations above to $\Sigma_\fin$ and $\tSigma_0$.
\begin{itemize}
    \item
The computation for $\Sigma_\fin$ is similar to the above, with an important addition: when computing the complexity of critical points with index at most $k = \eps d$, one imposes the constraint $\indi\{\mathrm{i}[\Hess \, \hR(\btheta)] \leq k\}$ in eq.~\eqref{eq:KR_1}. This results in a constraint $\indi\{\mathrm{i}[H(\bU)] \leq k\}$ in the expectation in eq.~\eqref{eq:KR_2}. 
    We give here a heuristic account for how this additional constraint influences the computation.
    Recalling that the law of $(g(\bU), H(\bU))$ only depends on the empirical law $\nu$ of $(\bu_i)_{i=1}^n$, we separate two cases:
    \begin{itemize}
        \item If $\supp \mu_{\alpha}[\nu] \subseteq [t(\nu), \infty)$, then we simply upper bound: 
        \begin{align}\label{eq:ub_index}
            \EE_{\{\bz_i\}}\left[\left|\det H(\bU)\right| \indi\{{\rm i}[H(\bU)] \leq k\} \middle| g (\bU) = 0\right] \leq \EE_{\{\bz_i\}}\left[\left|\det H(\bU)\right| \middle| g (\bU) = 0\right].
        \end{align}
        \item If $\supp \mu_{\alpha}[\nu] \not\subseteq [t(\nu), \infty)$, then the asymptotic spectral law of $H(\bU)$ puts positive mass on $(-\infty, 0)$.
        For sufficiently small $\eps > 0$, the event $\indi\{{\rm i}[H(\bU)] \leq \eps d\}$ has then probability $\exp\{-\Theta(d^2)\}$ since it requires moving the whole spectral density, while the large deviations of the spectral density has rate $d^2$.
    \end{itemize}
    The conclusion is that the computation of the complexity $\Sigma_\fin(Q, B)$ yields an upper bound which is again eq.~\eqref{eq:Sigma_TC_a}, with the additional constraint $\supp \mu_\alpha[\nu] \subseteq [t(\nu), \infty)$ in the supremum over $\nu$.
This is the result obtained in~\cite{asgari2025local}, where it is used as an upper bound on the complexity of local minima.

    \myskip 
    Finally, we notice that
    eq.~\eqref{eq:ub_index} should be asymptotically tight for any $k = \eps d$ as $d \to \infty$. 
    Indeed since the asymptotic spectral law of $H(\bU)$ is supported on $[0, \infty)$ in this case, the probability that $H(\bU)$ has a number of negative eigenvalues greater than $\eps d$ has probability $e^{-\Theta(d^2)}$.
    In a nutshell, we counted critical points whose Hessian's asymptotic spectral density is non-negatively supported: since the asymptotic spectral density is insensitive to perturbations in the Hessian with rank sub-extensive in $d$, one effectively counts all sub-extensive-index saddle points.
    
    \item Regarding $\tSigma_0$, the condition $(\btheta^\star)^\perp \Hess \hR(\btheta) \btheta^\star \geq 0$ can be written as a function of the empirical law $\nu$ as well, as it reads:
    \begin{align*}
       &\frac{1}{n} \sum_{i=1}^n \frac{(y^\star_i - q y_i)^2}{1-q^2} \partial_1^2 \ell(y_i, y_i^\star) - \frac{1}{n}\sum_{i=1}^n y_i \partial_1 \ell(y_i, y_i^\star) \\ 
       &= 
       \int \nu(\rd y, \rd y^\star) \left[\partial_1^2 \ell(y, y^\star) \frac{(y^\star - q y)^2}{1-q^2} - y \partial_1 \ell(y, y^\star)\right] 
       \geq 0.
    \end{align*}
    With respect to $\Sigma_\fin$,
    this simply accounts for an additional linear constraint on $\nu$ in the variational principle.
\end{itemize}

\myskip 
\textbf{Final results --}
The arguments above form the sketch of the derivation of asymptotic formulas, which we present in detail in Section~\ref{subsec:main_th_kr}.
In particular, notice that the formulas obtained are of the form
\begin{align*}
    \Sigma(Q, B) = \sup_{q \in Q} \sup_{e \in B} \Sigma(q, e). 
\end{align*}
In what follows, we therefore present asymptotic formulas for the functions $\tSigma_0(q, e), \Sigma_\fin(q, e)$ and $\Sigma_\TC(q, e)$.

\subsection{Main theoretical results from the Kac-Rice approach}\label{subsec:main_th_kr}

\subsubsection{Asymptotic formulas involving a supremum over probability measures}\label{subsubsec:var_formulas}

We state first the asymptotic formulas (as $n, d \to \infty$) of the complexities defined above, 
obtained through the derivation sketched in Section~\ref{subsec:kac_rice_sketch}, and which are natural transpositions and extensions of the results of~\cite{maillard2020landscape,asgari2025local}.
We state them for the general loss function of eq.~\eqref{eq:loss}: 
\begin{align*}
    \hR(\btheta) = \frac{1}{n} \sum_{i=1}^{n} \ell(\bx_i \cdot \btheta, \bx_i \cdot \btheta^\star).
\end{align*}

\myskip
\textbf{The total number of critical points --}
For completeness, let us recall eq. (\ref{eq:Sigma_TC_a}):
\begin{align}\label{eq:Sigma_TC}
    \nonumber
    \Sigma_\TC(q, e) &= \frac{1 + \log \alpha}{2} + \frac{1}{2} \log (1-q^2) \\ 
    &\hspace{10pt}+ \sup_{\nu \in \mcM_\ell(q, e)}\left[-\frac{1}{2} \log \int_{\bbR^2} \nu(\rd \bu) A(\bu)
    + \kappa_{\alpha}(\nu) - \alpha H(\nu | \mu_q)\right].
\end{align}
Recall that $\mcM_\ell(q, e)$ is the set of probability measures on $\bbR^2$ such that $\int \nu(\rd \bu) \ell(\bu) = e$ and 
$\int \nu(\rd \bu) c_q(\bu) = 0$, with $\bu \coloneqq (y, y^\star)$.
Moreover, we defined  $\mu_q$ in eq.~\eqref{eq:def_muq}, and $A, c_q$ in eq.~\eqref{eq:def_Ac}.
Finally, we have
\begin{align}\label{eq:def_kappa_t}
\begin{dcases}
    \kappa_{\alpha}(\nu) &\coloneqq \int \mu_{\alpha}[\nu](\rd w) \log |w - t(\nu)|, \\
    t(\nu) &\coloneqq \int \nu(\rd y, \rd y^\star) \, y \, \partial_1 \ell(y, y^\star).
\end{dcases}
\end{align}
And recall that $\mu_{\alpha}[\nu]$ is the asymptotic spectral distribution of $\bz \bD \bz^\T / n$, with $\bz \in \bbR^{d \times n}$ with i.i.d.\ $\mcN(0,1)$ elements, and 
$\bD = \Diag(\{ \partial^2_1 \ell(\bu_i)\}_{i=1}^n)$, for $\{\bu_i\}_{i=1}^n \iid \nu$.
Importantly, $\mu_{\alpha}[\nu]$ can be characterized straightforwardly via its Stieltjes transform~\cite{marchenko1967distribution,silverstein1995empirical}, see Appendix~\ref{subsec_app:Hessian_spectrum}.

\myskip 
\textbf{Local minima --}
Equipped with the definitions above, we can generalize the formula obtained for $\Sigma_\tot(q, e)$ to the other types of complexities we defined in Section~\ref{subsubsec:complexities}.
As discussed above, we do not provide a sharp result for the annealed complexity of local minima, but rather an upper bound $\tSigma_0(q, e)$.
It reads:
\begin{align}\label{eq:Sigma_0}
    \nonumber
    \tSigma_0(q, e) &= \frac{1 + \log \alpha}{2} + \frac{1}{2} \log (1-q^2) \\ 
    &\hspace{10pt}+ \sup_{\nu \in \mcM^\0_\ell(q, e)}\left[-\frac{1}{2} \log \int_{\bbR^2} \nu(\rd \bu) A(\bu)
    + \kappa_{\alpha}(\nu) - \alpha H(\nu | \mu_q)\right].
\end{align}
Here:
\begin{align}\label{eq:def_Mell_0}
    \mcM_\ell^\0(q, e) \coloneqq \{\nu \in \mcM(q, e) \, : \, &\supp(\mu_{\alpha}[\nu]) \subseteq [t(\nu), +\infty) \textrm{ and }\\ 
    & 
    \nonumber
    \int \nu(\rd y, \rd y^\star) \left[\partial_1^2 \ell(y, y^\star) \frac{(y^\star - q y)^2}{1-q^2} - y \partial_1 \ell(y, y^\star)\right] \geq 0\}.
\end{align}

\myskip 
\textbf{Saddles of sub-extensive index --}
For saddles of sub-extensive index on the other hand, we do reach an exact asymptotic formula for their annealed complexity.
It is almost identical to eq.~\eqref{eq:Sigma_0}:
\begin{align}\label{eq:Sigma_fin}
    \nonumber
    \Sigma_\fin(q, e) &= \frac{1 + \log \alpha}{2} + \frac{1}{2} \log (1-q^2) \\ 
    &\hspace{10pt}+ \sup_{\nu \in \mcM^\fin_\ell(q, e)}\left[-\frac{1}{2} \log \int_{\bbR^2} \nu(\rd \bu) A(\bu)
    + \kappa_{\alpha}(\nu) - \alpha H(\nu | \mu_q)\right].
\end{align}
The only difference with respect to eq.~\eqref{eq:Sigma_0} is the lack of the last constraint on $\nu$ in the variational principle, as:
\begin{align*}
    \mcM_\ell^\fin(q, e) \coloneqq \{\nu \in \mcM(q, e) \, : \, &\supp(\mu_{\alpha}[\nu]) \subseteq [t(\nu), +\infty) \}.
\end{align*}

\myskip 
\textbf{Connection with previous works --}
Eq.~\eqref{eq:Sigma_TC} is exactly the result of~\cite{maillard2020landscape}, up to a reparametrization. 
On the other hand, the main result of~\cite{asgari2025local} reduces to the formula we present for sub-extensive-index critical points $\Sigma_\fin(q, e)$, see eq.~\eqref{eq:Sigma_fin}. The formula for $\tSigma_0(q, e)$ on the other hand is new: while a simple variation over the one for $\Sigma_\fin(q, e)$, we will see that the additional constraint proves crucial in order to establish topological trivialization from our results.

\myskip 
\textbf{Characterization of critical points --}
Crucially, the derivation in Section~\ref{subsec:kac_rice_sketch} shows that the variational formulas above provide access not only to the complexity, but also to other relevant statistical properties of the critical points. In particular, the measure $\nu$ achieving the supremum corresponds to the prediction for the joint law of $(y, y^\star)$ for local minima, sub-extensive-index saddles, or generic critical points as predicted by the annealed Kac–Rice formalism. 
 Moreover, the asymptotic spectral density of the Hessian at these points can also be characterized: it is given by $\mu_{\alpha}[\nu]$, up to a shift $t(\nu)$ induced by the spherical constraint.
We emphasize that these predictions should be regarded as approximations, since we consider only the annealed complexity (i.e.\ its first moment), and our computation for local minima is an upper bound.

\subsubsection{Simplifying the problem: asymptotic formulas involving only scalar parameters}

The variational formulation over probability measure can be greatly simplified and recast 
into a variational principle involving only scalar parameters. The detailed derivation of these results is deferred to Appendix~\ref{subsec_app:functional_to_scalar}. Here we just present the final outcome. 

\myskip
\textbf{Local minima --}
We start with local minima, i.e.\ eq.~\eqref{eq:Sigma_0}.
To slightly lighten the statement of the formula, let us define three additional functions first (recall that $\bu = (y, y^\star)$):
\begin{align}\label{eq:def_FtK}
    \begin{dcases}
        F(\bu) &\coloneqq \partial_1^2 \ell(y, y^\star), \\
        t(\bu) &\coloneqq y \, \partial_1 \ell(y, y^\star), \\
        K_q(\bu) &\coloneqq \partial_1^2 \ell(y, y^\star) \frac{(y^\star - q y)^2}{1-q^2} - y \partial_1 \ell(y, y^\star).
    \end{dcases}
\end{align}
Recall also the definition of $\mu_q$ in eq.~\eqref{eq:def_muq}.
The formula for $\tSigma_0(q, e)$ can be recast into the following variational principle:
\begin{align}
    \label{eq:Sigma_0_scalar}
    &\tSigma_0(q, e) 
    = \frac{-1 + (1-2\alpha) \log \alpha}{2} + \frac{1}{2} \log(1-q^2) \\ 
     \nonumber
     &+ \sup_{\substack{g > 0 \\ A > 0}} 
     \inf_{\substack{\lambda_c, \lambda_e, \lambda_t, \lambda_A \in \bbR \\ \lambda_h, \lambda_\star \geq 0}}
    \Bigg\{-\frac{1}{2} \log A + \alpha(\lambda_A A + \lambda_e e)
    - \log g - \frac{\lambda_t}{g} + \lambda_h \\
    &\nonumber
    + \alpha \log \int_{B_g} \mu_q(\rd \bu) \, e^{- \lambda_c c_q(\bu) - \lambda_A A(\bu) - \lambda_e \ell(\bu) + \log(\alpha + F(\bu) g)
    - \frac{g+\lambda_t}{\alpha} t(\bu) + \lambda_t \frac{F(\bu)}{\alpha + F(\bu) g}
    - \lambda_h\left(\frac{F(\bu) g}{\alpha + F(\bu) g}\right)^2 
    + \frac{\lambda_\star}{\alpha} K_q(\bu)
    } \Bigg\}.
\end{align}
Here $B_g \coloneqq \{\bu \in \bbR^2 \, : \, \alpha + g F(\bu) > 0\}$.
Some important remarks are in order:
\begin{enumerate}
    \item While eq.~\eqref{eq:Sigma_0_scalar} appears involved, the different Lagrange multipliers and parameters appearing have natural interpretations, which we discuss in Appendix~\ref{subsec_app:functional_to_scalar}.
    \item Notice that the integral in the last term might be infinite for some values of the parameters and Lagrange multipliers. Moreover, as we detail in the derivation, if $B_g \neq \bbR^2$ (i.e.\ if there is $\bu$ such that $\alpha + g F(\bu) = 0$) then 
    we must have either $\lambda_h > 0$ or $\lambda_h = 0$ and $\lambda_t \geq 0$:
    this ensures in particular that the integrand is always going to $0$ at the boundary of $B_g$.
    \item The inner infimum over the set of Lagrange multipliers in eq.~\eqref{eq:Sigma_0_scalar} is convex.
    While we can not guarantee that the outer $\sup$ over $(g, A)$ admits a unique global maximum, in our numerical applications (see Section~\ref{subsec:exploring_landscape}) we never encountered the existence of several local maxima to this functional, even though we varied the initial conditions.
\end{enumerate}
As we discussed above, this formula also provides an estimate for 
the asymptotic empirical law $\nu_0$ of $(\bx_i \cdot \btheta, \bx_i \cdot \btheta^\star)_{i=1}^n$, for $\btheta$ uniformly sampled from the set of critical points 
satisfying the properties in eq.~\eqref{eq:def_tSigma0}.
This prediction reads:
\begin{align}\label{eq:nu0}
    &\nu_0(\rd \bu) \\ 
    \nonumber &= \frac{\mu_q(\rd \bu) e^{- \lambda_c c_q(\bu) - \lambda_A A(\bu) - \lambda_e \ell(\bu) + \log(\alpha + F(\bu) g)
    - \frac{g+\lambda_t}{\alpha} t(\bu) + \lambda_t \frac{F(\bu)}{\alpha + F(\bu) g}
    - \lambda_h\left(\frac{F(\bu) g}{\alpha + F(\bu) g}\right)^2 
    + \frac{\lambda_\star}{\alpha} K_q(\bu)
    }}
    {\int_{B_g} \mu_q(\rd \bu) e^{- \lambda_c c_q(\bu) - \lambda_A A(\bu) - \lambda_e \ell(\bu) + \log(\alpha + F(\bu) g)
    - \frac{g+\lambda_t}{\alpha} t(\bu) + \lambda_t \frac{F(\bu)}{\alpha + F(\bu) g}
    - \lambda_h\left(\frac{F(\bu) g}{\alpha + F(\bu) g}\right)^2 
    + \frac{\lambda_\star}{\alpha} K_q(\bu)
    }}.
\end{align}
Here, all parameters and Lagrange multipliers are taken at their respective optimum in eq.~\eqref{eq:Sigma_0_scalar}.
Moreover, the asymptotic spectral density at the Hessian at a typical critical point satisfying eq.~\eqref{eq:def_tSigma0} is given by 
$\rho_0(\rd w)$, which satisfies for any function $f$:
\begin{align}\label{eq:rho0}
    \int \rho_0(\rd w) f(w) =  \int \mu_{\alpha}[\nu_0](\rd w) f(w - t(\nu_0)),
\end{align}

\myskip 
\textbf{Saddles of sub-extensive index --}
Given the similarity of eqs.~\eqref{eq:Sigma_fin} and \eqref{eq:Sigma_0}, 
we expect and find that $\Sigma_{\fin.}(q, e)$ satisfies essentially eq.~\eqref{eq:Sigma_0_scalar}, 
up to removing the Lagrange multiplier $\lambda_\star$:
\begin{align}
    \label{eq:Sigma_fin_scalar}
    &\Sigma_\fin(q, e) 
    = \frac{-1 + (1-2\alpha) \log \alpha}{2} + \frac{1}{2} \log(1-q^2) \\ 
     \nonumber
     &+ \sup_{\substack{g > 0 \\ A > 0}} 
     \inf_{\substack{\lambda_c, \lambda_e, \lambda_t, \lambda_A \in \bbR \\ \lambda_h \geq 0}}
    \Bigg\{-\frac{1}{2} \log A + \alpha(\lambda_A A + \lambda_e e)
    - \log g - \frac{\lambda_t}{g} + \lambda_h \\
    &\nonumber
    + \alpha \log \int_{B_g} \mu_q(\rd \bu) \, e^{- \lambda_c c_q(\bu) - \lambda_A A(\bu) - \lambda_e \ell(\bu) + \log(\alpha + F(\bu) g)
    - \frac{g+\lambda_t}{\alpha} t(\bu) + \lambda_t \frac{F(\bu)}{\alpha + F(\bu) g}
    - \lambda_h\left(\frac{F(\bu) g}{\alpha + F(\bu) g}\right)^2 
    } \Bigg\}.
\end{align}
One can directly transpose the predictions of eq.~\eqref{eq:nu0} to the law $\nu_\fin$ of the labels $(y_i, y_i^\star)_{i=1}^n$ at a typical sub-extensive-index saddle by setting $\lambda_\star = 0$, 
and the one of eq.~\eqref{eq:rho0} to the spectral density of the Hessian $\rho_\fin$.

\myskip
\textbf{Critical points --}
We finally consider generic critical points.
In this setting, we follow the heuristic approach suggested in~\cite{maillard2020landscape}. We obtain an asymptotic formula that is similar to the ones above, 
with a few key differences.
\begin{align}\label{eq:Sigma_TC_scalar}
    \Sigma_\TC(q, e) &= \frac{-1 + (1- 2\alpha)\log \alpha}{2} + \frac{1}{2} \log(1-q^2) 
    + \extr_{\substack{A > 0, g \in \bbC_+ \\ 
    \lambda_A, \lambda_e, \lambda_c \in \bbR}}
    \Bigg[- \frac{1}{2} \log A - \log |g| + \eps g_i \\ 
    \nonumber
    &+ \alpha (\lambda_A A + \lambda_e e) 
    + \alpha \log \int_{\bbR^2} \mu_q(\rd \bu) e^{- \lambda_c c_q(\bu) - \lambda_A A(\bu) - \lambda_e \ell(\bu) + \log |\alpha + F(\bu) g| - \frac{g_r}{\alpha} t(\bu)}\Bigg].
\end{align}
Here $\bbC_+ \coloneqq \{z \in \bbC \, : \, \Im(z) > 0\}$ we denoted $g = g_r + i g_i$, and we fixed a given small $\eps \ll 1$ (eq.~\eqref{eq:Sigma_TC_scalar} is understood in the limit $\eps \downarrow 0$).
An important difference with respect to eqs.~\eqref{eq:Sigma_0_scalar} and~\eqref{eq:Sigma_fin_scalar} is that the extremum is not given 
as a $\sup\inf$. Rather, one should find the saddle point of the functional that maximizes the complexity.
A second difference is that the integral over $\bu$ is now made over the whole $\bbR^2$, as the integrand is well-defined for all $\bu \in \bbR^2$.
Obtaining the predictions for $(\nu_\tot, \rho_\tot)$ from eq.~\eqref{eq:Sigma_TC_scalar} is again immediate: e.g. we have 
\begin{align}\label{eq:nu_tot}
    \nu_\tot(\rd \bu)
    &= \frac{\mu_q(\rd \bu) e^{- \lambda_c c_q(\bu) - \lambda_A A(\bu) - \lambda_e \ell(\bu) + \log |\alpha + F(\bu) g| - \frac{g_r}{\alpha} t(\bu)}}
    {\int_{\bbR^2} \mu_q(\rd \bu) \, e^{- \lambda_c c_q(\bu) - \lambda_A A(\bu) - \lambda_e \ell(\bu) + \log |\alpha + F(\bu) g| - \frac{g_r}{\alpha} t(\bu)}}.
\end{align}

\myskip 
\textbf{Numerical solutions --}
The relatively simple form of eqs.~\eqref{eq:Sigma_0_scalar},\eqref{eq:Sigma_fin_scalar} and~\eqref{eq:Sigma_TC_scalar} suggest natural algorithms to evaluate these complexity functions. In Appendix~\ref{subsec_app:scalar_to_alg} we detail the practical implementations we used for their evaluation: we use in particular the remark above on the convexity of the inner minimization problem. A complete implementation in Python using adaptive quadrature routines accelerated by Numba's just-in-time (JIT) compilation~\cite{lam2015numba} will be provided in a Github repository.

\myskip 
\textbf{Final remarks --}
In what follows, we will sometimes remove the dependency of the complexity functions on $(q, e)$, and write e.g.\ $\tSigma_0(q)$ or $\Sigma_\tot$. In general, the value of the overlap $q$ will always be fixed and precised. On the other hand, if the value of the energy/loss $e$ is not given, we are considering the maximal complexity across possible values of $e$: equivalently, this is achieved by fixing the Lagrange multiplier $\lambda_e = 0$  in the variational formulas written above.


\subsection{The BBP-Kac-Rice method, and the instability of local minima}\label{subsec:bbp_theory}
The Kac–Rice analysis provides access to the asymptotic eigenvalue density of the Hessian at the critical points under consideration. However, to relate the landscape properties to training dynamics, such as gradient flow, it is crucial to determine whether critical points---particularly local minima---may develop a descending direction, i.e. an instability, aligned with the signal as $\alpha$ varies.
Such a descending direction would be a low-rank change in the Hessian, which were not considered by the Kac–Rice method. Instead, we address this problem with random matrix theory, determining the presence of such an outlier by the so-called ``BBP'' transition~\cite{baik2005phase,benaych2011eigenvalues}.

\myskip
Concretely, the spherical Hessian reads here (see Section~\ref{subsec:kac_rice_sketch}):
\begin{align}\label{eq:Hessian_spherical_bbp}
    \Hess = \frac{1}{n} \sum_{i=1}^n \partial_1^2 \ell(y_i, y_i^\star) (P_\btheta^\perp \bx_i) (P_\btheta^\perp \bx_i)^\T - \left(\frac{1}{n} \sum_{i=1}^n y_i \partial_1 \ell(y_i, y_i^\star)\right) P_{\btheta}^\perp,
\end{align}
where $P_\btheta^\perp$ is the orthogonal projection on $\{\btheta\}^\perp$. 

\myskip 
Crucially, one can develop an analytical theory for whether the Hessian given by eq.~\eqref{eq:Hessian_spherical_bbp} develops an informative outlier, for \emph{any estimator} $\nu$ of the asymptotic joint law of the labels. $(y_i, y_i^\star)$. This extends a previous analysis of~\cite{bonnaire2025role}, which only considered the Hessian at the equator ($q = \btheta \cdot \btheta^\star = 0$).
The derivation and main equations are provided in Appendix~\ref{sec_app:bbp}.

\myskip 
\textbf{The BBP-KR method --} 
\begin{figure}[!t]
    \centering
    \includegraphics[width=0.65\textwidth]{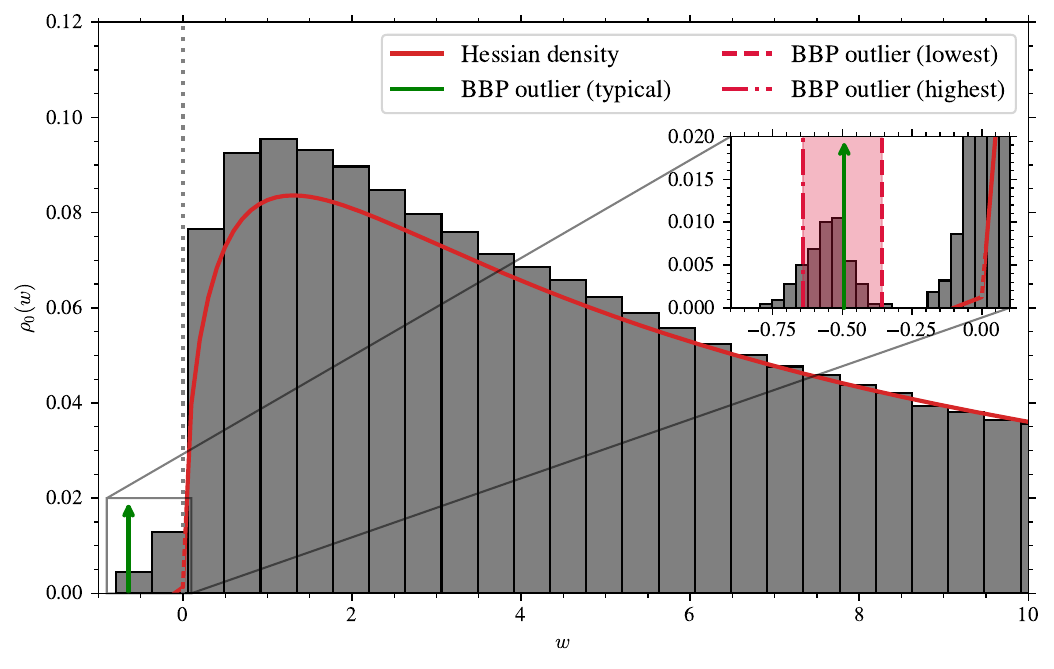}
    \caption{
        \centering
        In the phase retrieval problem (see eq.~\eqref{eq:def_ell_a}), with $a = 0.01$, $q = 0.0$, and $\alpha = 6.5$, we show the annealed Kac-Rice prediction for the Hessian density of typical-energy minima, see eq.~\eqref{eq:rho0}. 
        The predicted complexity is $\tSigma_0(q) \simeq 7.10^{-3} > 0$ (see also Fig.~\ref{fig:phase_diagram} below).
        The green arrow represents the position of the negative ``BBP'' outlier in the spectrum predicted by our theory for this Hessian density, see eq.~\eqref{eq:outlier}. 
        The red-shaded area delimitates the prediction for the location of the outlier for all minima with positive complexity, from the highest-energy to lowest-energy ones.
        The histograms correspond to numerical results obtained using finite-size gradient descent simulations, see Section~\ref{subsec:numerical_setting} for more details.
        %
        \label{fig:BBP_KR_description}
    }
\end{figure}
The above point is the basis of the ``BBP-Kac-Rice'' prediction: 
namely, the BBP instability for the Hessian in eq.~\eqref{eq:Hessian_spherical_bbp} can be derived from the label law $\nu$ predicted by the Kac–Rice analysis, e.g.\ eq.~\eqref{eq:nu0}.
Strikingly, as we will see in concrete examples, this analysis predicts that a BBP instability of the Hessian can occur while the Kac-Rice method still predicts a positive complexity, and hence the presence of exponentially many local minima.
We illustrate this phenomenon in Fig.~\ref{fig:BBP_KR_description} for phase retrieval, where we plot the eigenvalue density obtained from the Kac–Rice bound on the complexity of minima for $a = 0.01$, $q = 0.0$, and $\alpha = 6.5$. While the bound suggests here a positive complexity of minima, the BBP–KR analysis instead reveals the presence of a negative outlier. The associated eigenvector is correlated with the signal, indicating that the putative minima are in fact unstable towards the signal.\footnote{The fact that $\boldsymbol{\theta}^*$ is not a descending direction is not in contradiction with the presence of a negative BBP eigenvalue, since the corresponding eigenvector has only a finite overlap with the signal $\boldsymbol{\theta}^*$.}
%
\subsection{Applications: exploring the landscape of phase retrieval}\label{subsec:exploring_landscape}

We now present the applications of our asymptotic formulas to the problem of phase retrieval, with the loss of eq.~\eqref{eq:def_ell_a}:
\begin{align*}
    \ell_a(y, y^\star) \coloneqq \frac{(y^2 - [y^\star]^2)}{a + [y^\star]^2}.
\end{align*}
From now on, we fix $a = 0.01$: nevertheless, our results can be straightforwardly extended to any value of $a > 0$: as examples, we extend some of the forthcoming results to 
$a \in \{0.1, 1.0\}$ in Appendix~\ref{sec_app:more_kac_rice}.
\\ 
An interesting property of this loss is that the denominator ensures that the second derivative $\partial_1^2 \ell$ is bounded from below: 
in turns this ensures that the Hessian's asymptotic spectral density is always bounded from below. 

\myskip 
In the following we present explicit results obtained from our variational formulas applied to the phase retrieval problem. 
As we aim to keep a concise presentation in the main text,
we defer some additional results and a deeper exploration to Appendix~\ref{sec_app:more_kac_rice}, 

\subsubsection{Complexity of local minima}

\begin{figure}[!t]
    \centering
    \includegraphics[width=1.0\textwidth]{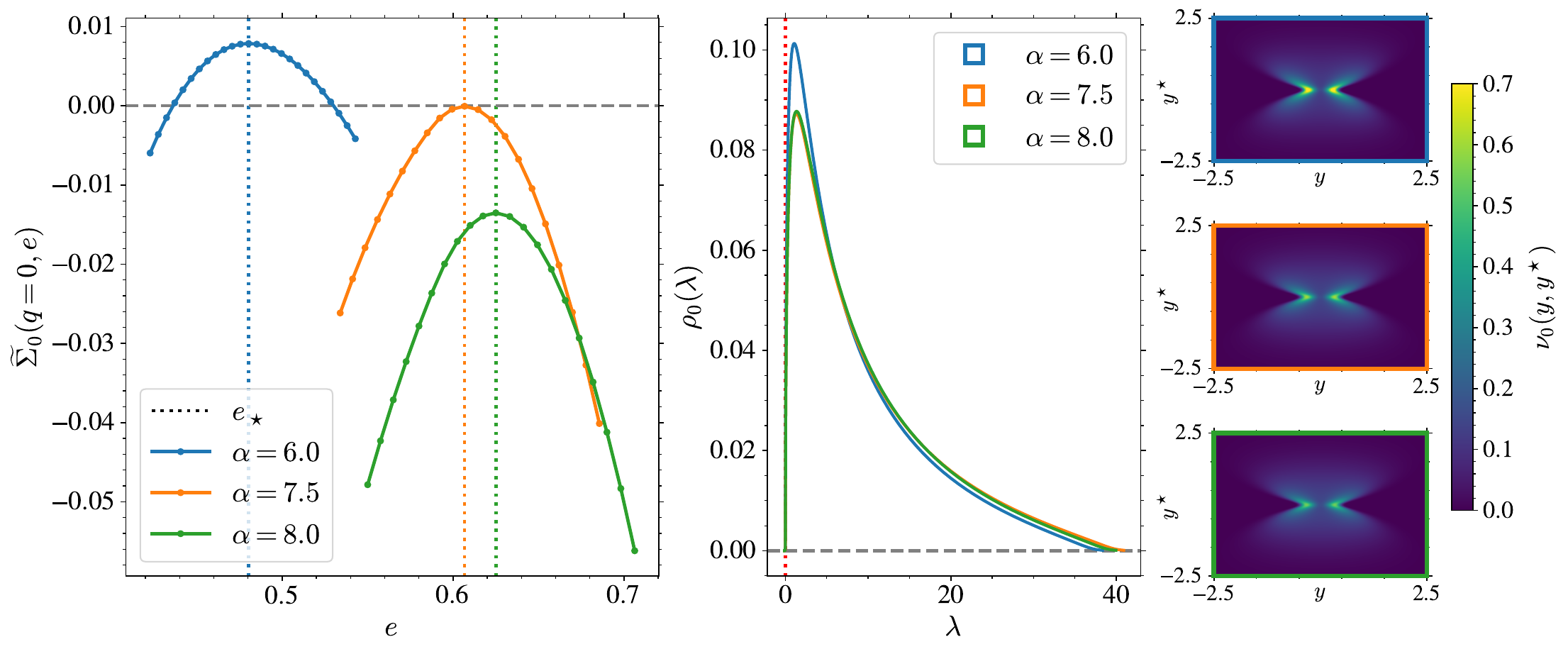}
    \caption{
        \centering
        For $a = 0.01$, $q = 0$, and three values $\alpha = 6.0 < \alpha_{\triv}$, $\alpha =  7.5 \simeq \alpha_{\triv}$ and $\alpha = 8.0 > \alpha_{\triv}$, 
        we show:
        (left) the complexity $\tSigma_0(q=0, e)$ for different values of $e$ around the maximal complexity,
        (middle) the density of the Hessian at local minima of typical energy (at $e_\star \coloneqq \argmax \tSigma_0(q = 0, e)$), 
        and (right) the corresponding law $\nu(y, y^\star)$.
        The results are obtained by solving the optimization problem of eq.~\eqref{eq:Sigma_0_scalar}.
        \label{fig:minima_kac_rice}
    }
\end{figure}

We first consider in Fig.~\ref{fig:minima_kac_rice} our upper bound on the complexity of local minima, i.e.\ the function $\tSigma_0(q, e)$.
We plot this complexity function as a function of $e$, for $q = 0$ and different values of $\alpha = n/d$, and show corresponding predictions for the label's distribution $\nu_0(y, y^\star)$ and the Hessian spectral density $\rho_0(w)$.
Several important comments can be made on these results: 
\begin{itemize}
    \item For small $\alpha$, there is a band of energy $[e_{\min}, e_{\max}]$ where local minima are located\footnote{Except the trivial global minimum $\btheta = \btheta^\star$, located exactly at $e = 0$.}, i.e.\ where $\tSigma_0(q, e) > 0$. 
    The most numerous of them (with maximal complexity) have an energy $e_\star$. 
    \item As $\alpha$ increase, the size of this band diminishes: it eventually disappears at $\alpha = \alpha_\triv \simeq 7.5$.
    \item The typical Hessian of local minima varies relatively little close to this transition. In particular, it always remains \emph{marginally stable} (the bulk of eigenvalues touches $0$). This is an interesting and remarkable properties of the dominant minima. It is possible that stable ones exist but are less numerous (sub-dominant). We leave this issue for a future work, where a more refined computation, targeting also sub-dominant minima, will be performed.  
    \item 
    We emphasize that a random point $\btheta \in \mcS^{d-1}$, such as the one from which the gradient flow dynamics start from, would produce a standard Gaussian law for $\nu(y, y^\star)$: in contrast, 
    the right plots of Fig.~\ref{fig:minima_kac_rice} show that, for $\btheta$ a local minimum, the estimated $y_i = \btheta \cdot \bx_i$ manage to align themselves very well with the ground-truth labels $y^\star_i = \btheta^\star \cdot \bx_i$ despite $\btheta$ not aligning itself at all with $\btheta^\star$ (recall $q = 0$). 
\end{itemize}

\subsubsection{Complexity of critical points and saddles of sub-extensive index}

\begin{figure}[!t]
    \centering
    \includegraphics[width=1.0\textwidth]{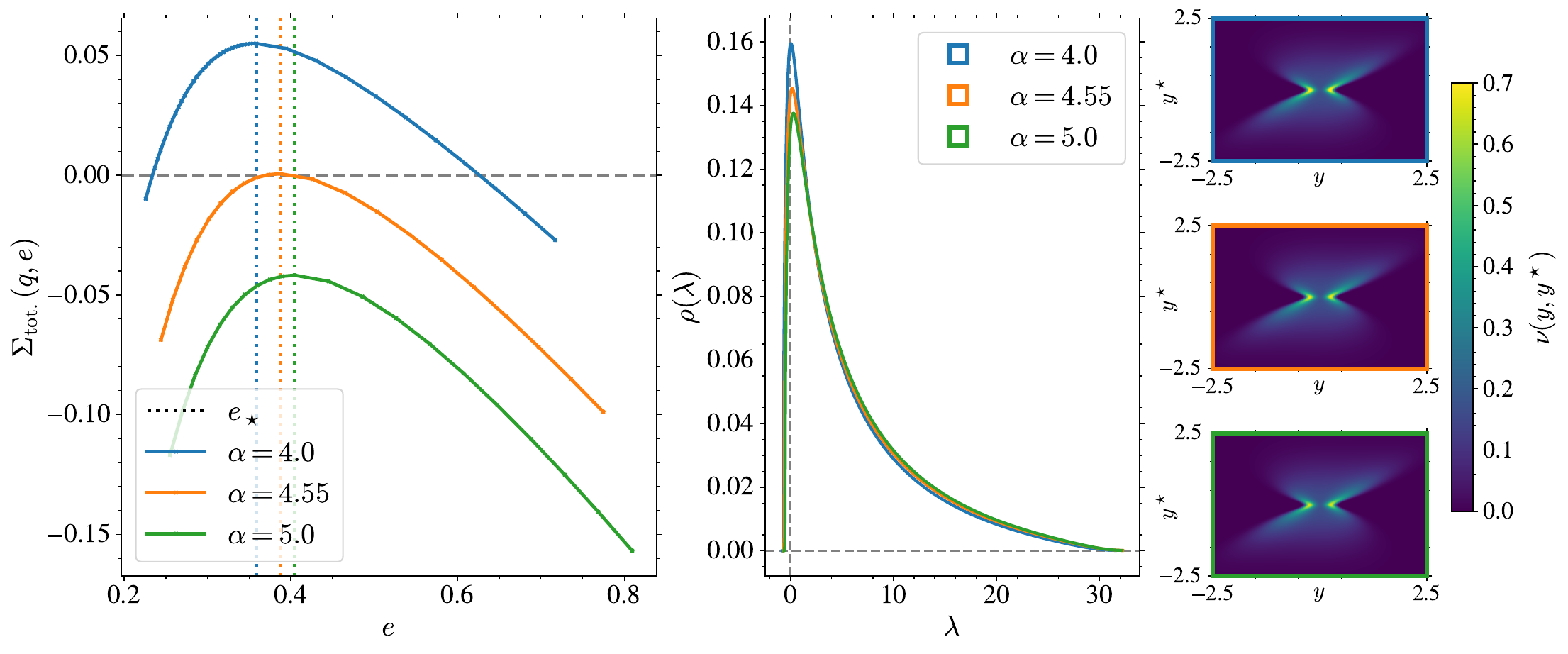}
    \caption{
        \centering
        For $a = 0.01$, $q = 0.4$, and three values $\alpha = 4.0 < \alpha_{\triv}$, $\alpha =  4.55 \simeq \alpha_{\triv}$ and $\alpha = 5.0 > \alpha_{\triv}$, 
        we show:
        (left) the complexity $\Sigma_\TC(q, e)$ for different values of $e$ around the maximal complexity,
        (middle) the density of the Hessian at critical points of typical energy (at $e_\star \coloneqq \argmax \Sigma_\TC(q, e)$), 
        and (right) the corresponding law $\nu(y, y^\star)$.
        The results are obtained by solving the optimization problem of eq.~\eqref{eq:Sigma_TC_scalar}.
        \label{fig:critical_points_kac_rice}
    }
\end{figure}

In Fig.~\ref{fig:critical_points_kac_rice}, we show a similar analysis to Fig.~\ref{fig:minima_kac_rice}, but for generic critical points. 
We focus on a value $q = 0.4$ which shows a trivialization transition. Contrary to the case of minima, we don't find such transition for $q=0$ at any finite $\alpha$ -- a point that we will discuss further in   Section~\ref{subsubsec:global_phase_diagram}.
    As shown in Fig. \ref{fig:critical_points_kac_rice}, the critical points exhibit features similar to those previously discussed for local minima. In particular, they concentrate within a well-defined energy band, and the law $\nu(y, y^\star)$ displays a clear alignment between the estimated labels and the ground-truth ones.
The Hessian is no longer positive definite: its spectrum contains a small but nonzero fraction of negative eigenvalues, both below and above the trivialization threshold in $\alpha$.

\myskip 
\textbf{Sub-extensive-index saddles --}
We do not discuss  $\Sigma_{\mathrm{fin}}$ here in order to keep the exposition concise. Its behavior closely parallels that of the critical points; in particular, there is no trivialization of sub-extensive-index saddles at $q = 0$ for any finite value of $\alpha$. This point will be illustrated and discussed in Section~\ref{subsubsec:global_phase_diagram} (see Fig.~\ref{fig:phase_diagram}).

\subsubsection{Properties of typical-energy, low-energy, and high-energy minima}

\begin{figure}[!t]
    \centering
    \includegraphics[width=\linewidth]{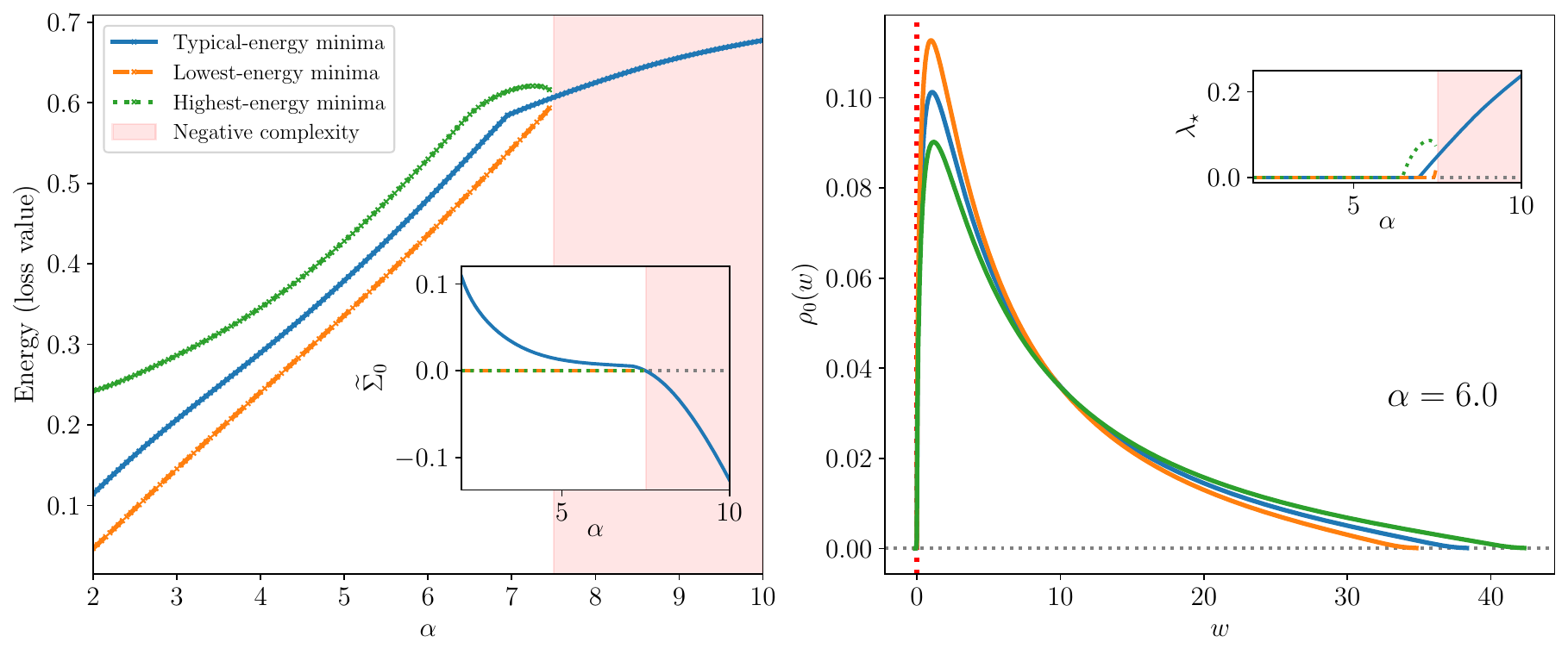}
    \caption{
        \centering
        For $a = 0.01$, $q = 0.0$, we show (left) the energy of the typical, lowest, and highest-energy minima as a function of $\alpha$. The inset shows the value of the complexity $\tilde \Sigma_0$ (notice that highest and lowest-energy minima are always at zero complexity by definition).
        In the red region 
        the complexity of local minima is negative.
        (Right) The spectral density $\rho_0(w)$ of the Hessian at these minima, for $\alpha = 6.0$. We show in dotted red the line $w = 0$.
        In the inset, we show the Lagrange multiplier $\lambda_\star$ as a function of $\alpha$.
        The results are obtained by solving the Kac-Rice formula of eq.~\eqref{eq:Sigma_0_scalar}.
        \label{fig:minima_low_high_energy}
    }
\end{figure}

As we saw above in Fig.~\ref{fig:minima_kac_rice}, for fixed $q \in [0,1)$ and $\alpha < \alpha_\triv(q)$ the annealed Kac-Rice formalism predicts the existence of a band of local minima.
In Fig.~\ref{fig:minima_low_high_energy} we show the width of this band for $q = 0$, as well as the Hessian density of highest and lowest-energy minima.
We find that the width of this energy band diminishes as $\alpha$ increases. Around a value of $\alpha_\star \simeq 7.0$, one can observe a ``kink'' in the value of the typical energy: it corresponds to the activation of the last constraint in eq.~\eqref{eq:def_Mell_0}, and is apparent from the onset of $\lambda_\star$ away from $0$ (see top-right inset, and the discussion in Appendix~\ref{sec_app:theory_kac_rice}). For $\alpha \lesssim  7.0$ local minima are more numerous than other saddles of sub-extensive index. Instead for $\alpha \simeq 7.0$  the complexity of local minima becomes strictly smaller and then vanishes at $\alpha_\triv$, whereas the complexity of saddles of finite index remains positive \emph{ at any finite $\alpha > 0$}. 
Nevertheless, we remind that these claims concern the \emph{annealed} complexity (i.e.\ the first moment of the number of critical points): while a negative annealed complexity implies the absence of the corresponding critical points, a positive annealed complexity is not sufficient to argue their existence with high probability, and should rather seen as an indication of this existence. Finally, we also show in Fig.~\ref{fig:minima_low_high_energy} the Hessian density for the lowest-energy, typical-energy, and highest-energy minima.
As we will see in Section~\ref{subsubsec:global_phase_diagram}, lowest-energy minima are the ``most stable'' under the BBP instability described in Section~\ref{subsec:bbp_theory}. This was also shown on a specific example in Fig.~\ref{fig:BBP_KR_description}, and we give further numerical simulations in Appendix~\ref{subsec_app:bbp_kr}.

\subsubsection{Global phase diagrams}\label{subsubsec:global_phase_diagram}

\begin{figure}[!t]
  \begin{subfigure}{\linewidth}
    \centering
    \includegraphics[width=\linewidth]{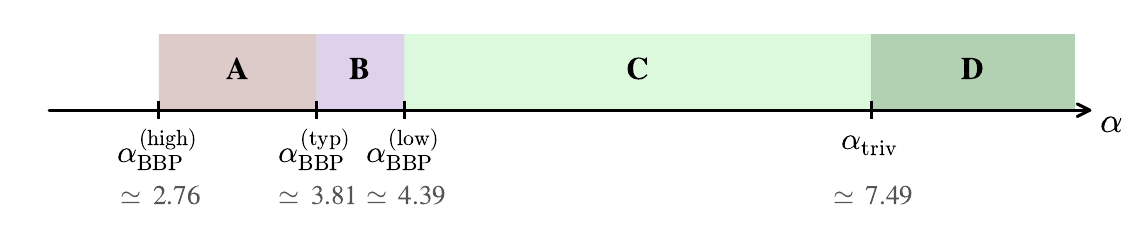}
   \caption{
        For $q = 0$, the BBP thresholds for highest, typical, and lowest-energy minima, and the trivialization of local minima.
        As $\alpha$ increases, the BBP-instability propagates through the energy landscape: 
        it first emerges in high-energy minima $(\text{A})$, then spreads to the 
        majority of minima $(\text{B})$, and eventually encompasses even the 
        lowest-energy states $(\text{C})$. Finally, in $(\text{D})$, the Kac-Rice upper bound implies the absence of any local minima.
   }
   \label{subfig:phase_diagram_q_0}
  \end{subfigure}

  \medskip 

  \begin{subfigure}{\linewidth}
    \centering
    \includegraphics[width=.83\linewidth]{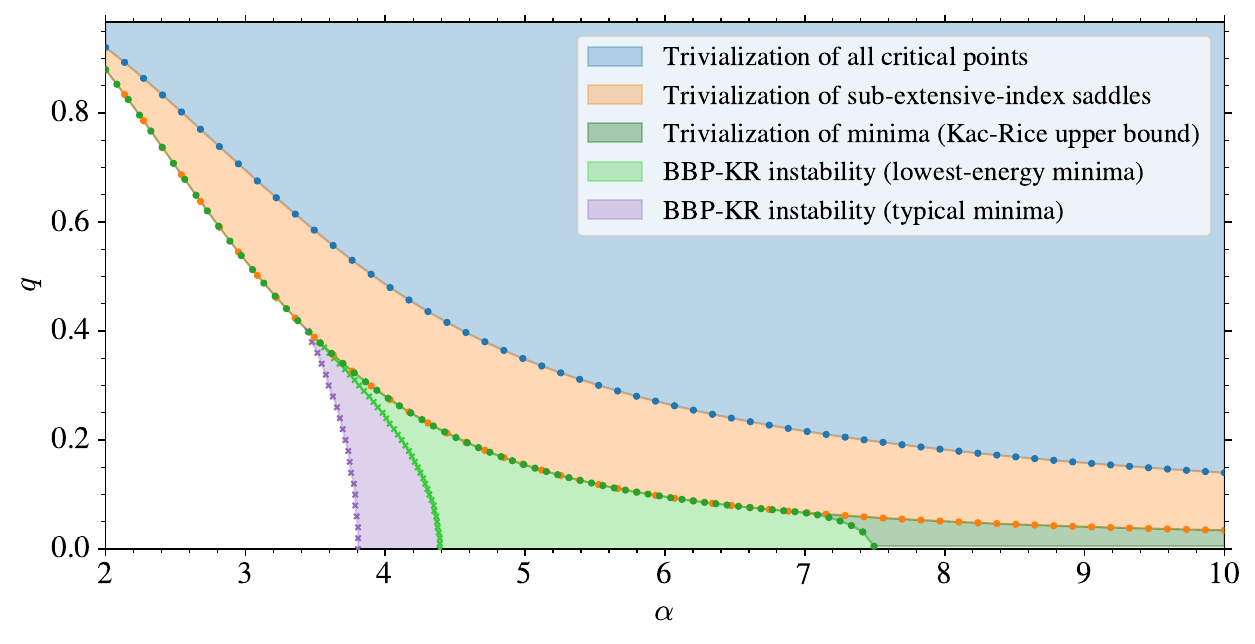}
   \caption{
        A complete phase diagram in the $(\alpha, q)$ plane, showing regions of trivializations of the landscape, where our Kac-Rice analysis implies the absence of 
        critical points, of saddles of sub-extensive index, and of local minima. We also draw the BBP-instability regions for typical-energy minima, and for the lowest-energy minima:
        in the light green region, the annealed Kac-Rice formalism predicts that all minima have a BBP instability towards the signal, before our upper bound shows the absence of minima in the dark green area.
   }
    \label{subfig:phase_diagram}
  \end{subfigure}
    \caption{
        Phase diagram predicted by the Kac-Rice method, for $a = 0.01$.
        \label{fig:phase_diagram}
    }
\end{figure}

Finally, we show in Fig.~\ref{fig:phase_diagram} the complete topological phase diagram of the phase retrieval problem (with $a = 0.01$).
In Fig.~\ref{subfig:phase_diagram_q_0} we showcase the different thresholds for $q = 0$ as a function of $\alpha$. We see that the BBP-KR instability appears first for high-energy minima at $\alpha = \alpha_\BBP^{(\rm high)} \simeq 2.76$, before propagating to all local minima, and finally lowest-energy minima at $\alpha = \alpha_\BBP^{(\rm low)} \simeq 4.39$. This is intuitive given that the low energy minima are expected to be more stable than the highest energy ones. 

At this value of $\alpha$, while our Kac-Rice formalism still predicts a positive annealed complexity of local minima, the BBP-KR analysis allows us to unveil that minima undergo a BBP instability towards the signal. The Kac-Rice formalism is unable to detect this transition because it does not focus on low-rank contributions to the Hessian. Only for a larger value $\alpha_\triv \simeq 7.49$, the Kac-Rice formalism predicts a negative complexity. With some abuse of notations we still call this the ``trivialization threshold'', but emphasize that the true trivialization transition takes place before, when all minima become unstable due to the BBP instability. An example of this phenomenon was already illustrated in Fig.~\ref{fig:BBP_KR_description}, which was made for $q = 0$ and $\alpha = 6.5 \in (\alpha_\BBP, \alpha_\triv)$.

\myskip
In Fig.~\ref{subfig:phase_diagram} we show the same thresholds in the $(\alpha, q)$ plane. We observe several trivialization regions, corresponding either to a BBP instability or to the disappearance of some types of critical points.
Notably, there is a blue region where the landscape does not possess any critical point (besides the trivial minimum $\btheta = \btheta^\star$ at $q = 1$).
Moreover, we emphasize that the orange region (the trivialization of finite-index saddles) never reaches $q = 0$ for any finite $\alpha > 0$: we therefore expect that the computations of~\cite{maillard2020landscape,asgari2025local}, based on the characterization of $\Sigma_\fin$, would not yield a finite trivialization threshold $\alpha_\triv$ in this problem, 
while it is known that local minima disappear at a finite $\alpha$~\cite{cai2021globalII}.

\subsubsection{Final remarks}\label{subsubsec:final_remarks}

We conclude this analysis with a couple of remarks.
First of all, while we discussed some illustrative results above, we present in Appendix~\ref{sec_app:more_kac_rice} many other results and a more systematic exploration of the empirical risk landscape using the Kac-Rice formalism.
We also discuss there a possible limitation of our computations: as we discussed in Section~\ref{subsec_app:functional_to_scalar}, in general we can not guarantee uniqueness of a solution to the scalar variational problems of eqs.~\eqref{eq:Sigma_0_scalar},\eqref{eq:Sigma_fin_scalar} and~\eqref{eq:Sigma_TC_scalar}.
In practice, we never found that there existed different local maxima to the complexity formulas of local minima and sub-extensive-index saddles. However, we did exhibit in one limited regime the coexistence of two saddle points for the complexity formula of critical points (eq.~\eqref{eq:Sigma_TC_scalar}), and it is associated to a first-order phase transition that we discuss in Appendix~\ref{subsec_app:phase_transition_total_complexity}.
A more systematic study of such phenomena, and a proof of the uniqueness of the solutions to the complexity formulas for local minima and sub-extensive-index saddles, are interesting lines of work for future research.

%% file: sections/simulations_dynamics.tex
The purpose of this section is to compare the predictions of the annealed Kac–Rice analysis with the properties of minima obtained from numerical simulations of finite-dimensional gradient-descent (GD) dynamics. We first describe the numerical setup in Section~\ref{subsec:numerical_setting}. We then present a detailed statistical analysis of the GD minima, focusing on their energies, Hessian spectra, and second derivatives of the loss. All of these observables show a remarkably close agreement with the annealed theoretical predictions. Finally, in Section~\ref{subsec:numerical_phase_diagram}, we confront the theoretical phase diagram with the empirical outcomes of GD in phase retrieval across varying values of $q$ and $\alpha$, thereby directly connecting landscape structure to the observed success or failure of GD.

\subsection{Settings of the experiments} \label{subsec:numerical_setting}

Given a dataset $\{\bx_i\}_{i=1}^n$ with $\bx_i \sim \mcN(0, \Id_d)$ and a signal vector $\btheta^\star \sim \mcN(0, \Id_d/d)$, we aim to minimize the empirical risk~\eqref{eq:loss} starting from an initialization $\btheta^{(0)}$, and using full-batch GD updates,
\begin{equation} \label{eq:GD_update}
    \btheta^{(t+1)} = \btheta^{(t)} - \eta \,\nabla \hR\!\left(\btheta^{(t)}\right),
\end{equation}
for $t\in\{0, 1, \dots, T-1\}$ with fixed learning rate $\eta$. Throughout this section, we consider the loss function defined in eq.~\eqref{eq:def_ell_a} and fix the normalization parameter to $a=0.01$. We are going to focus on two kinds of initialization: (1) uncorrelated with the signal: $\btheta\sim \mcN(0, \Id_d/d)$, (2) warm start: the component of $\btheta$ along $\btheta^*$ is fixed to $q_0$, and all the other components (in orthogonal directions) are i.i.d. Gaussian variables with mean zero and variance $1/d$. 


\myskip 
\textbf{Finite-dimension effects, and modified dynamics --} 
Our goal is to compare the Kac--Rice predictions, derived in the limit \(d \to \infty\), with numerical experiments representative of the high-dimensional behavior. In the phase retrieval problem, however, the gradient-descent dynamics~\eqref{eq:GD_update} is strongly affected by finite-dimensional effects. As shown in~\cite{bonnaire2025role}, convergence to the infinite-dimensional limit at finite but large times \(t\) requires, for certain values of \(\alpha\), dimensions that grow exponentially with \(t\). This phenomenon originates from a time-dependent BBP transition in the Hessian over a range of \(\alpha\): at early times, the Hessian exhibits a negative outlier eigenvalue whose eigenvector is aligned with the signal, while this outlier disappears at later times (still at times of order one with respect to \(d\)). In the infinite-dimensional limit, the initial overlap with the signal is of order \(1/\sqrt{d}\), which prevents this instability from growing over any finite time: the outlier disappears  before becoming relevant.

In finite dimensions, by contrast, the instability induces an overlap with the signal that grows exponentially in time, so that a timescale of order \(\log d\) is sufficient to generate an \(O(1)\) overlap. This mechanism is responsible for the strong finite-dimensional corrections. Since simulating dimensions that grow exponentially with \(t\) is computationally infeasible, \cite{bonnaire2025role} introduced a modified dynamics that suppresses these corrections while preserving the same infinite-dimensional asymptotic limit. For completeness, we now explain this modified dynamics that we adapt to also probe minima at $q>0$. First, we draw $\btheta^{(0)}$ at random conditioned on having an initial overlap $q_0$ with the signal, i.e. $\btheta^{(0)}\cdot\btheta^\star=q_0$. Then, we perform $t_\mathrm{C}$ full-batch GD steps with a constraint enforcing that the overlap $\btheta^{(t)} \cdot \btheta^\star$ remains equal to $q_0$ during the descent. In practice, after each update, we project $\btheta^{(t)}$ back onto the manifold $\{\btheta \in \mcS^{d-1} \, : \, \btheta \cdot \btheta^\star = q_0\}$ for all $t<t_\mathrm{C}$ by writing
\begin{equation}
    \btheta^{(t)} \leftarrow q_0 \btheta^\star + \sqrt{1-q_0^2} \frac{\btheta_\perp^{(t)}}{\lVert \btheta_\perp^{(t)}\rVert_2^2}, \qquad \btheta_\perp^{(t)} = \btheta^{(t)} - \left( \btheta^{(t)} - \btheta^\star \right)\btheta^\star.
\end{equation}
The constrained iterate $\btheta^{(t_\mathrm{C})}$ therefore corresponds to a low-energy state with an overlap of exactly $q_0$ with the signal that has near-zero gradients in all directions except $\btheta^\star$. In a last stage, we use $\btheta^{(t_\mathrm{C})}$ to initialize a standard (unconstrained) GD dynamics. This procedure has the effect of considerably shifting the success rates to larger values of $\alpha$, more representative of the large $d$ limit \cite{bonnaire2025role}. To make the comparison between the dynamics and the predictions meaningful, we focus on minima reached at the end of the dynamics falling in a narrow latitude band around a fixed $q$, i.e. satisfying $q(T) \in \left[q - 0.05, q+0.05\right]$. 

\myskip 
\textbf{Hyperparemeters and success criterion --} In all our experiments, we fix the dimension to $d=512$ and the learning rate to $\eta=2\times 10^{-4}$. The considered loss function is the one from eq.~\eqref{eq:def_ell_a} with $a=0.01$. Additional comparison of the dynamics with the Kac-Rice predictions are shown for $a=1$ in Appendix~\ref{sec_app:dynamics_general_a}. The maximum number of GD iterations in both phases is set to $t_\mathrm{C}=60{,}000$ and $T=12{,}000 \log_2(d)$. We classify a run as successful if the final absolute overlap $|\btheta^{(T)} \cdot \btheta^\star| > 0.99$, and as a failure otherwise. For each set of $(\alpha, q_0)$, we draw between $1{,}000$ and $2{,}400$ independant realizations (by resampling both the dataset and the initial condition $\btheta^{(0)}$) to compute means and error bars. We made sure that $\eta$ is sufficiently small to approximate gradient flow by observing convergence of the success rate at fixed $(\alpha, q)$ as $\eta$ is decreased.

\subsection{Properties of minima: Kac-Rice predictions vs numerical experiments} \label{subsec:numerical_comparison}

\begin{figure}
    \centering
    \includegraphics[width=\textwidth]{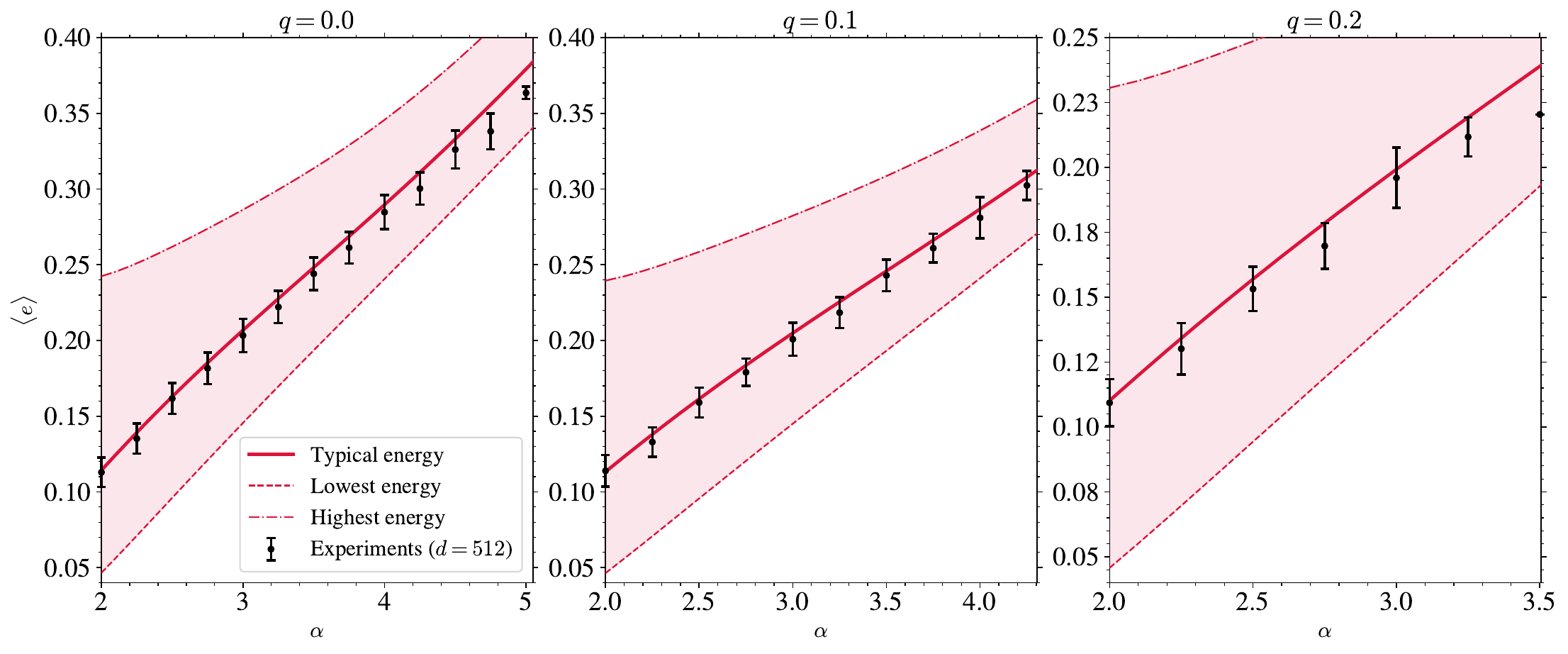}
    \caption{
        \centering Evolution of the average energy $\langle e \rangle$ of the minima for $a=0.01$ with $\alpha$ at fixed (left) $q=0.0$, (middle) $q=0.1$, and (right) $q=0.2$ obtained from the experiments (in black) and the band of energy predicted by the annealed Kac-Rice (in shaded red).
        \label{fig:energies_a0.01_KR_exp}
    }
\end{figure}

We begin our evaluation of the annealed Kac-Rice predictions by comparing the energies of minima reached by our finite-size dynamics at $d=512$. Figure~\ref{fig:energies_a0.01_KR_exp} reports, for $q \in \{0.0, 0.1, 0.2\}$, the average energies of minima $\langle e \rangle$ reached by our numerical procedure as a function of $\alpha$, together with the energy band predicted analytically by Kac-Rice. We observe an excellent agreement across the entire explored range of $\alpha$ where the energies of GD minima lie inside the band predicted by the Kac-Rice method for lowest, typical and highest minima. The energies of GD minima even closely align with the typical energies predicted by Kac-Rice. This observation supports the view that, when GD fails, its dynamics do not converge to exotic, low-probability minima, but rather get trapped in typical regions of the loss landscape.

\begin{figure}
  \begin{subfigure}{\linewidth}
    \centering
    \includegraphics[width=.9\linewidth]{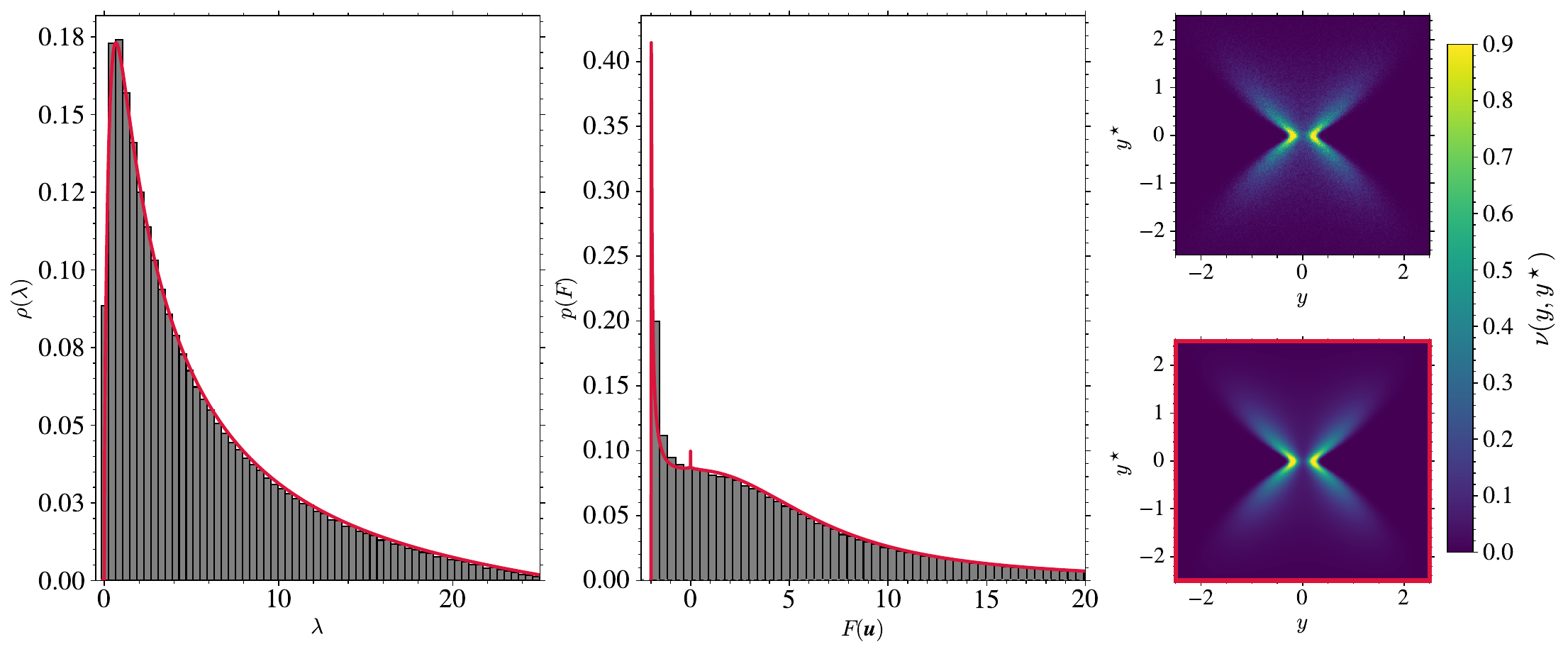}
   \caption{
        For $\alpha = 3.5$.
   }
  \end{subfigure}

  \medskip

  \begin{subfigure}{\linewidth}
    \centering
    \includegraphics[width=.9\linewidth]{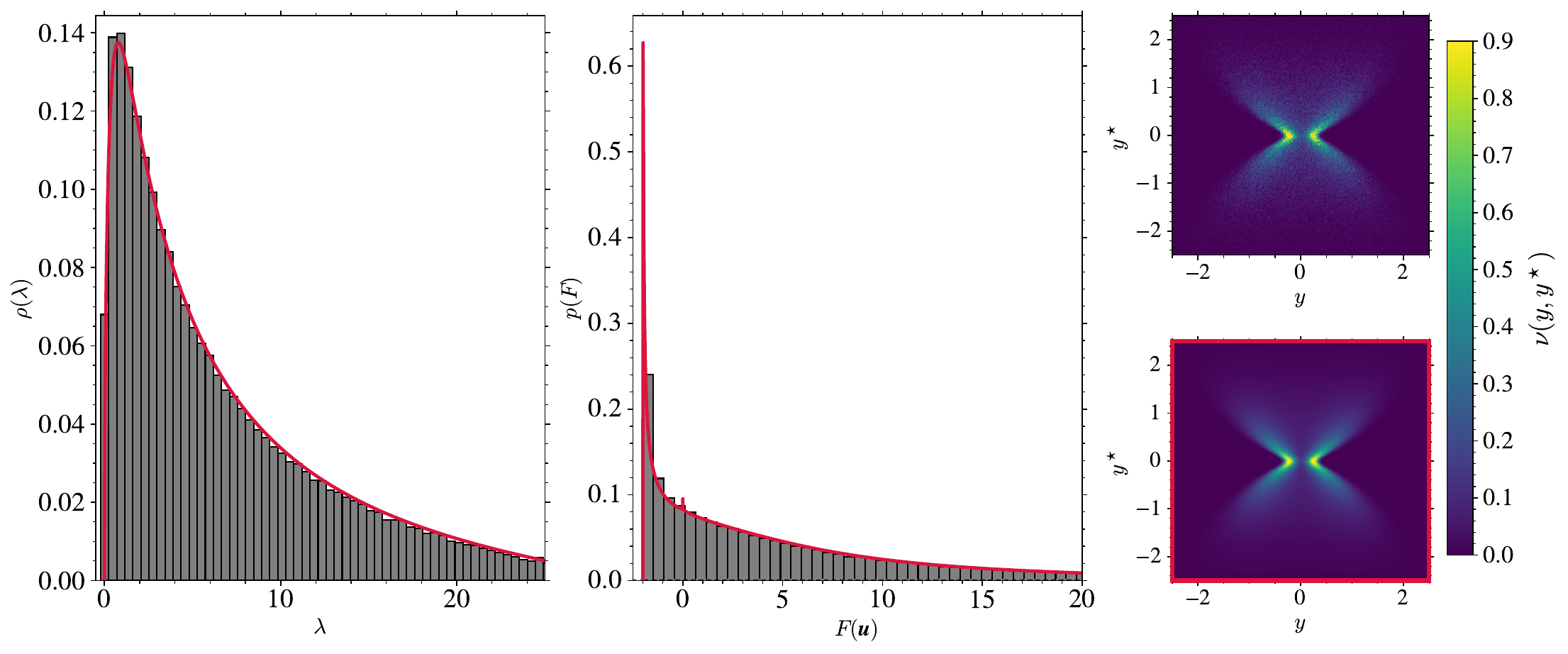}
   \caption{
        For $\alpha=4.5$.
   }
  \end{subfigure}
  
    \caption{
        Comparisons of the predicted Kac-Rice properties for the minima at $q=0.0$ and typical energy $e_\star$ and the empirical minima found by the gradient descent dynamics at $d=512$ and $a=0.01$. We compare: (left) the eigenvalue distribution of the Hessian $\rho(\lambda)$, (middle) the distribution of the Hessian weights $F(\bu)=\partial^2_1 \ell(y,y^\star)$, and (right) the joint label distributions $\nu(y, y^\star)$. In red are the Kac-Rice predictions described in Section~\ref{sec:kac_rice}.
        \label{fig:comparison_dynamics_H_nu_PF}
    }
\end{figure}

\myskip
We then probe more properties of these minima in Fig.~\ref{fig:comparison_dynamics_H_nu_PF} by comparing, for $q=0$ and $\alpha\in\{3.5, 4.5\}$, three observables evaluated for typical minima at energy $e_\star$. From left to right, we show: (i) the spectral density of Hessian eigenvalues at converged minima; (ii) the distribution of $F(\bu)=\partial^2_1 \ell(y,y^\star)$; and (iii) the joint distribution $\nu(y, y^\star)$ of predicted and true labels, which is the central object from which all the quantities are derived.
For all three observables, the theoretical predictions (in red) match the empirical distributions remarkably well, not only around their typical values, but also in the tails. A minor discrepancy is visible for the singular behavior of $F(\bu)$ at zero which is not resolved by our finite-sample analysis.
The theoretical and experimental distributions $\nu(y,y^\star)$ are visually indistinguishable, both being similarly sharp and exhibiting the same structure. We emphasize that this match is highly non-trivial as a generic random configuration at fixed $q$ would produce a jointly Gaussian distribution by standard high-dimensional arguments, whereas the minima selected by the dynamics display a clear non-Gaussian shape. Observing this non-Gaussian structure prediction matching cleanly the finite-$d$ experiments is a particularly strong validation of the ability of our theoretical method to capture properties of minima of the loss landscape. Overall, these comparisons indicate that, even at a moderate dimension $d=512$, the annealed Kac-Rice predictions provide an accurate picture of the minima properties in the energy landscape as a function of $(\alpha, q)$.


\subsection{Phase diagram from Kac-Rice vs gradient descent algorithmic thresholds} \label{subsec:numerical_phase_diagram}

\begin{figure}
  \begin{subfigure}{\linewidth}
    \centering
    \includegraphics[width=\linewidth]{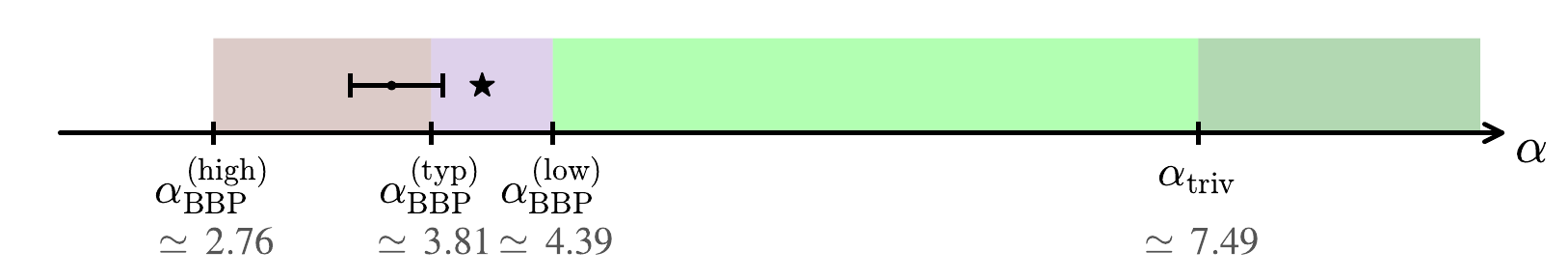}
   \caption{
        Same as Fig.~\ref{subfig:phase_diagram_q_0} with the experimental success transition for $q=0.0$ (error bar). The star corresponds to the finite-size analysis of the same loss function in \cite{bonnaire2025role}.
   }
   \label{fig:phase_diagram_exp_a}
  \end{subfigure}

  \medskip

  \begin{subfigure}{\linewidth}
    \centering
    \includegraphics[width=.83\linewidth]{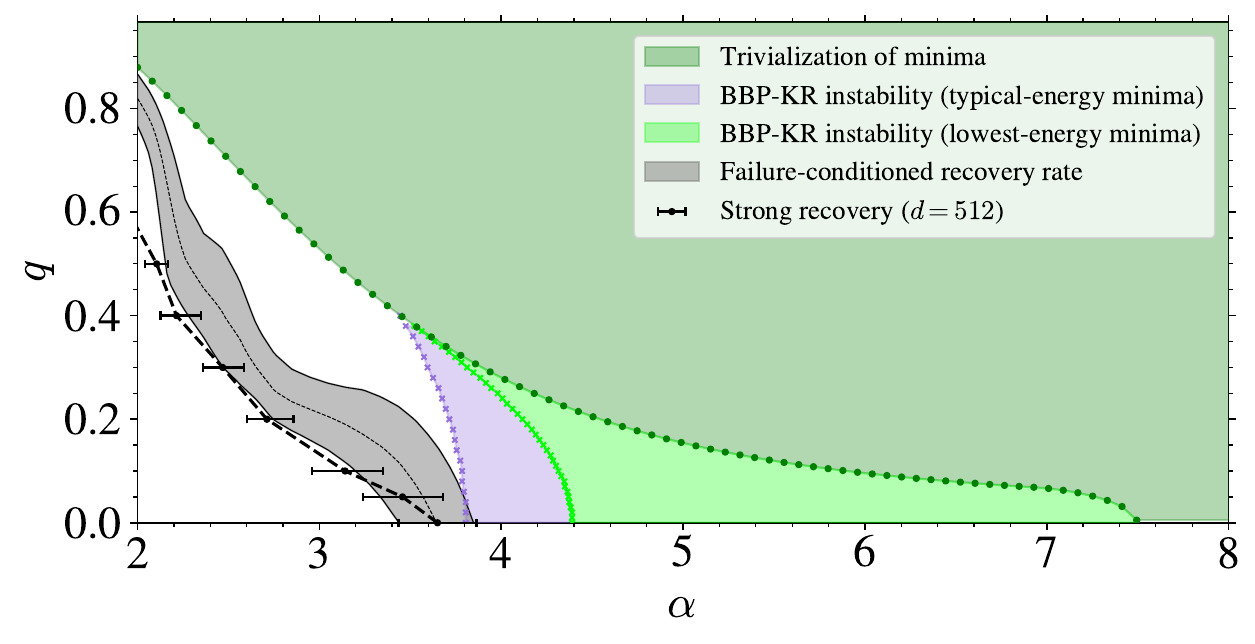}
   \caption{
        Minima-only phase diagram from Fig.~\ref{fig:phase_diagram} with the strong recovery rates obtained from our numerical experiments ($d=512$). Success is observed empirically in the region consistently below the predicted BBP and trivialization transitions at all $(\alpha, q)$.
   }
   \label{fig:phase_diagram_exp_b}
  \end{subfigure}
  
    \caption{
        Minima phase diagram predicted by the Kac-Rice method for $a=0.01$ along with the experimental results.
        \label{fig:phase_diagram_exp}
    }
\end{figure}

Following \cite{Mannelli2019,Mannelli2020b,bonnaire2025role}, the success or failure of gradient descent in phase retrieval, and more generally in generalized linear models, can be related to the landscape properties discussed in Section~\ref{subsubsec:global_phase_diagram}. In particular, the algorithmic recovery threshold\footnote{Strictly speaking, the Kac-Rice BBP method should be informative on the {\it weak} recovery threshold for random initial condition ($q_0=0$). However, in our simulations we always find strong recovery when recovery is achieved. Therefore, we do not distinguish between weak and strong recovery in what follows. For warm starts ($q_0>0$), our numerical results show that, at sufficiently small $\alpha$, gradient descent is trapped in minima with overlap larger than $q_0$. As $\alpha$ increases, we observe a transition toward strong recovery. We identify this transition with the one predicted by the BBP–Kac–Rice analysis.} is conjectured to coincide with the BBP instability of the local minima located on the equator ($q=0$), which trap the dynamics at low signal-to-noise ratio.
Since we observe a band of marginally stable minima, it remains unclear which ones of them constitute the “bad minima” responsible for trapping the dynamics. Addressing this question would require characterizing their basins of attraction — a highly challenging and, at present, open problem. A plausible working hypothesis is that the dynamically relevant minima are the typical ones, i.e., those that are most numerous\footnote{If the basins of attraction were independent of the energy of the minima, this would indeed be the case.}.\\
The aim of this section is therefore to compare the algorithmic thresholds found by numerical experiments on GD with the landscape phase diagrams obtained by the Kac-Rice approach. 
For convenience, Fig.~\ref{fig:phase_diagram_exp} reproduces the phase diagrams restricted to minima and overlays the finite-size experimental results. Our Kac-Rice analysis predicts that the signal-to-noise ratio $\alpha_\mathrm{triv}$ at which minima disappear from the landscape is larger than the BBP threshold $\alpha_\mathrm{BBP}^\mathrm{(typ)}$ at which typical-energy minima start to acquire an instability aligned with the signal $\btheta^\star$. One then expects three dynamical regimes in the $d\to\infty$ limit:
\begin{itemize}
    \item For $\alpha < \alpha_\mathrm{BBP}^\mathrm{(typ)}$, the typical minima are (marginally) stable and GD starting from a random initial condition is trapped into them at long times,
    \item For $\alpha \in \left[\alpha_\mathrm{BBP}^\mathrm{(typ)}, \alpha_\mathrm{triv}\right]$, these minima turn into saddles with a negative curvature towards $\btheta^\star$, enabling GD to escape and reach the signal in finite time,
    \item For $\alpha > \alpha_\mathrm{triv}$, all the minima disappear and the landscape become trivial to descend, so GD converges reliably to the signal.
\end{itemize}

The threshold values change if the initial condition has a finite overlap with the signal (warm start). 
Let us first focus on the case of an uncorrelated initial condition ($q=0$). Figure~\ref{fig:phase_diagram_exp_a} shows that the empirical success, shown as an error bar, is in very good agreement with the BBP-KR transition predicted for typical minima. The error bar is obtained by considering the $\alpha$ giving a success rate of $0.5\pm0.2$. It is also consistent with the finite-size analysis of \cite{bonnaire2025role}, reporting a BBP transition at $\alpha\approx4.0$ that we display as a star on the diagram. This highlights that GD can indeed succeed in a regime where an exponential number of minima are still present in the landscape, well before the trivialization threshold, as previously discussed in \cite{Mannelli2020b}.

\myskip
We extend this comparison to the full $(\alpha, q)$ plane in Fig.~\ref{fig:phase_diagram_exp_b}, completing the phase diagram from Fig.~\ref{subfig:phase_diagram} with the success-rate line extracted from GD experiments at $d=512$. We find that GD succeeds in the lower-left region of the diagram where Kac-Rice predicts the existence of stable spurious minima (including marginally stable ones), consistent with the idea that the practical success is primarily controlled by a BBP-like transition which is not visible using the Kac-Rice bound, but requires the BBP-Kac-Rice method we developed.
As $q$ increases away from 0, the success rate (dashed line and error bars) rapidly shifts to smaller values of $\alpha$ and progressively deviates from the theoretical BBP prediction for typical-energy minima. In particular, when $q_0 > 0$, we systematically observe that unsuccessful runs do not remain at their initial overlap: even when GD fails to reach the signal, $\langle q(T)\rangle_\mathrm{fail} > q_0$ for large-enough $\alpha$.
After a warm start at $q_0$, GD is expected to be trapped by minima with higher overlap before the instability towards the signal takes place. In consequence, in order to compare more fairly theory and numerical experiments, we report a second experimental line in which we use the failure-conditioned average $\langle q(T)\rangle_\mathrm{fail}$ in place of $q_0$ when assessing strong recovery. This correction shifts the empirical contour to larger $\alpha$, but a clear quantitative gap with the theoretical boundary remains. We interpret this discrepancy as likely due to (i) unresolved finite-size effects; and (ii) the need to go beyond the annealed Kac-Rice prediction and perform a quenched computation \cite{Ros2019,maillard2020landscape}. In particular, a quenched result is expected to shift the theoretical transition to lower $\alpha$ with respect to the annealed one.


%% file: sections/conclusion.tex
This work presents a systematic theoretical and numerical investigation of the geometry of non-convex empirical risk landscapes for generalized linear models, with an emphasis on phase retrieval. Combining the Kac-Rice method with random matrix theory, we derive tractable scalar variational formulas for the annealed complexities of local minima, sub-extensive-index saddles, and generic critical points, and we use them to build detailed phase diagrams for the phase retrieval landscape as a function of the sample complexity $\alpha = n/d$ and the overlap $q$ with the ground truth. We introduce a BBP--Kac-Rice method to further identify when local minima acquire a signal-aligned instability before the Kac-Rice trivialization threshold is reached, therefore providing a theoretical connection between landscape geometry and gradient-descent dynamics. These predictions are supported by extensive numerical experiments: the resulting energy bands, Hessian spectral properties, and label distributions are in excellent quantitative agreement with finite-size gradient-descent simulations at $d=512$, validating the practical accuracy of the formalism we developed.

\paragraph*{Limitations --} Several limitations of the present work deserve mention.
First, the complexities we compute are \emph{annealed} (i.e.\ based on the first moment $\EE[\#\{\text{critical points}\}]$) rather than \emph{quenched}: a positive annealed complexity does not guarantee the existence of exponentially many critical points with high probability, and whether the two notions coincide remains an open problem.
Second, we cannot guarantee in general the uniqueness of the solution to the scalar variational problems of eqs.~\eqref{eq:Sigma_0_scalar}--\eqref{eq:Sigma_TC_scalar}: while we never encountered multiple local maxima for the complexity of local minima or sub-extensive-index saddles, we did find coexisting solutions for $\Sigma_\TC$ in a limited regime, associated with a first-order phase transition (see Appendix~\ref{subsec_app:phase_transition_total_complexity}).
Third, our numerical implementation of the Kac--Rice formulas becomes unreliable as $q \to 1$, where the reference measure $\mu_q$ grows increasingly concentrated and the integrands develop sharp features.
Another difficulty arises on the dynamical side when considering large $q$s: at large warm-start overlaps, finite-size effects become harder to control, making it difficult to reliably estimate the large-dimensional algorithmic threshold in this regime.

\myskip
Finally, let us mention that it would be interesting to compare the results of our approach with the ones that can be obtained by the replica method, which has been often used in the past to perform landscape analysis. In an extended version of this work we shall complement the Kac--Rice analysis with a one-step replica-symmetry-breaking (1RSB) approach. The 1RSB method yields less accurate results, even at a qualitative level, for key observables such as the joint label distribution $\nu(y, y^\star)$ and the Hessian spectral density at local minima; we discuss possible explanations in the extended version.

\section*{Acknowledgements}
During the completion of this work we became aware of a concurrent work \cite{montanari2026topological} that also investigates the geometrical properties of non-convex empirical loss landscape of generalized linear models by the Kac-Rice method. A detailed comparison will be presented in an extended version of this work.
GB acknowledges support from the French government under the management of the Agence Nationale PR[AI]RIE-PSAI (ANR-23-IACL-0008).

%% file: sections/appendix/theory_kac_rice.tex


\subsection{Remainders on the Marchenko-Pastur equation}\label{subsec_app:Hessian_spectrum}

We recall here how to compute $\mu_\alpha[\nu]$ (see Section~\ref{subsubsec:var_formulas}), both analytically and with numerical procedures.
As we saw, it is (up to a global shift due to the spherical constraint) the asymptotic spectral density of the Hessian.
Recall that 
$\mu = \mu_{\alpha}[\nu]$ is the LSD of $\bz \bD \bz^\T / n$, with $\bz \in \bbR^{d \times n}$ an i.i.d.\ $\mcN(0,1)$ matrix, and 
$\bD = \Diag(\{F(\bu_i)\}_{i=1}^n)$ with $\{\bu_i\}_{i=1}^n \iid \nu$.

\myskip 
\textbf{Bulk's density --}
The Stieltjes transform $g(z) \coloneqq \int \mu(\rd t)  (t-z)^{-1}$ is given by the unique solution in $\bbC_+$ to the equation, 
for all $z \in \bbC_+$~\cite{marchenko1967distribution,silverstein1995empirical}:
\begin{align}\label{eq:stieltjes_sample_covariance}
    g(z) &= - \Bigg[z - \alpha \int \nu(\rd \bu) \frac{F(\bu)}{\alpha + g(z) F(\bu)} \Bigg]^{-1}. 
\end{align}
In practice, to compute $\mu(x)$ to solve eq.~\eqref{eq:stieltjes_sample_covariance} for $z = x + i \eps$ with $x \in \bbR$, $\eps > 0$ and $\eps \ll 1$. 
Using the Stieltjes-Perron inversion theorem, this indeed allows to get the density of $\mu$: 
\begin{align}\label{eq:stieltjes-perron}
    \frac{\rd \mu}{\rd x} &= \lim_{\eps \to 0} \frac{1}{\pi} \Im[g(x + i \eps)]. 
\end{align}
After evaluating the law of $X \coloneqq F(\bu)$ for $\bu \sim \nu$, one can
solve eq.~\eqref{eq:stieltjes_sample_covariance} numerically very efficiently by using a simple root solver, as explained e.g.\ in~\cite{couillet2022random}.
More precisely, we solve $G(g) = 0$, with 
\begin{align}\label{eq:def_G}
    G(g) \coloneqq g + \left(z - \alpha \int \nu(\rd \bu) \frac{F(\bu)}{\alpha + g(z) F(\bu)}\right)^{-1}.
\end{align}
Note as well that we 
can rewrite eq.~\eqref{eq:stieltjes_sample_covariance} in terms of the inverse $g^{-1}(s)$ of the Stieltjes transform:
\begin{align}\label{eq:inv_stieltjes_sample_covariance}
   g^{-1}(s) &= - \frac{1}{s} + \alpha \int \frac{F(\bu)}{\alpha + s F(\bu)} \nu(\rd \bu).
\end{align}

\myskip 
\textbf{Edge of the spectrum --}
In full generality, the support of the density can be computed by investigating regions where the right-hand side of eq.~\eqref{eq:inv_stieltjes_sample_covariance} is real and increasing.
Under mild assumptions (see~~\cite{silverstein1995analysis,lee2016tracy,couillet2022random} for more details), 
the left edge of the spectrum is given by the smallest positive solution to $[g^{-1}]'(s) = 0$, i.e.\ 
the unique solution to
\begin{align}\label{eq:gm1_prime}
    \alpha \int \Bigg(\frac{s F(\bu)}{\alpha + s F(\bu)}\Bigg)^2 \nu(\rd \bu) &= 1,
\end{align}
More generally, by~\cite[Theorems~4.1 and 4.2]{silverstein1995analysis}, we have that the Stieltjes transform of the left edge is the largest $S \geq 0$ such that 
both the following are true:
\begin{enumerate}[label=$( \roman* )$]
    \item If $X = F(\bu)$ for $\bu \sim \nu$, then $\supp(\alpha + S X) \subseteq (0, \infty)$. In particular the right-hand side of eq.~\eqref{eq:gm1_prime} is well-defined.
    \item 
    \begin{equation*}
        \alpha \int \Bigg(\frac{S F(\bu)}{\alpha + S F(\bu)}\Bigg)^2 \nu(\rd \bu) \leq 1.
    \end{equation*}
\end{enumerate}

\myskip
This yields then the left edge of the support as $x_{\min} = g^{-1}[S]$ given by eq.~\eqref{eq:inv_stieltjes_sample_covariance}.

\subsection{From functional to scalar variational formulas}\label{subsec_app:functional_to_scalar}


\subsubsection{Local minima and saddles of sub-extensive index}

We start from eq.~\eqref{eq:Sigma_0}. The derivation for $\Sigma_\fin$ follows exactly the same lines, removing a single constraint.
\begin{align*}
    \tSigma_0(q, e) &= \frac{1 + \log \alpha}{2} + \frac{1}{2} \log (1-q^2) \\ 
    &\hspace{10pt}+ \sup_{\nu \in \mcM^\0_\ell(q, e)}\left[-\frac{1}{2} \log \int_{\bbR^2} \nu(\rd \bu) A(\bu)
    + \kappa_{\alpha}(\nu) - \alpha H(\nu | \mu_q)\right].
\end{align*}
With some notations introduced in eq.~\eqref{eq:def_FtK}, we have 
\begin{align*}
    &\mcM_\ell^\0(q, e) \coloneqq \\
    &\{\nu \, : \, \EE_\nu[\ell(\bu)] = e, \, \,\EE_\nu[c_q(\bu)] = 0, \, \,  \supp(\mu_{\alpha}[\nu]) \subseteq [\EE_\nu[t(\bu)], +\infty) \textrm{ and }
     \EE_\nu[K_q(\bu)]\geq 0\}.
\end{align*}
Let $t(\nu) \coloneqq \EE_{\nu}[t(\bu)]$. 
We denote $g_\mu(z) \coloneqq \int \mu(\rd x) / (x - z)$ the Stieltjes transform of a probability measure $\mu$.
Importantly, one can show (cf.\ e.g.~\cite{maillard2020landscape}):
\begin{align}\label{eq:kappa_g_fixed}
    \kappa_{\alpha}(\nu) =  - \log g - t(\nu) g + \alpha \EE_{\nu} \log (\alpha + F(\bu) g) - 1 - \alpha \log \alpha,
\end{align}
with $g = g_{\mu_\alpha[\nu]}[t(\nu)] > 0$ (since $\nu \in \mcM_\ell^0(q, e)$).
Thus we get:
\begin{align}
    \label{eq:Sigma0_1}
    &\tSigma_0(q, e) = \frac{-1 + (1 - 2\alpha) \log \alpha}{2} + \frac{1}{2} \log (1-q^2) \\ 
    \nonumber
    &+ \sup_{\substack{g > 0 \\ A > 0}} \sup_{\substack{\nu \in \mcM^\0_\ell(q, e) \\ g_{\mu_\alpha[\nu]}(t(\nu)) = g \\ \EE_\nu[A(\bu)] = A}}\left[- \log g - \frac{1}{2} \log A - g \EE_\nu[t(\bu)] 
    + \alpha \EE_{\nu} \log[\alpha + gF(\bu)] - \alpha H(\nu | \mu_q)\right].
\end{align}
Using eq.~\eqref{eq:stieltjes_sample_covariance} and the characterization of the left edge of the support of $\mu_\alpha[\nu]$ which we recalled in Section~\ref{subsec_app:Hessian_spectrum}, for any $\nu$ and $g$ the condition
\begin{equation}\label{eq:def_C1}
    \mcC(\nu, g) \, : \, \left(\supp(\mu_{\alpha}[\nu]) \subseteq [t(\nu), +\infty) \textrm{  and  } g_{\mu_\alpha[\nu]}(t(\nu)) = g\right)
\end{equation}
can be recasted as the following three simultaneous conditions:
\begin{enumerate}[label=$( \roman* )$]
    \item  $\alpha + g F(\bu) \geq 0$, for all $u \in \supp(\nu)$.
    \item We have 
    \begin{equation}\label{eq:g_smaller_than_edge}
     \alpha \EE_{\nu} \left[\left(\frac{g F(\bu)}{\alpha + g F(\bu)}\right)^2\right] \leq 1.
    \end{equation}
    \item $g$ is the Stieltjes transform of $\mu_\alpha[\nu]$ in $t(\nu)$, i.e.:
    \begin{equation}\label{eq:stieltjes_t_g}
        -\frac{1}{g} - \EE_\nu[t(\bu)]+ \alpha \EE_\nu \left[\frac{F(\bu)}{\alpha + g F(\bu)}\right] = 0.
    \end{equation}
\end{enumerate}
The first condition constrains $\supp(\nu) \subseteq B_g \coloneqq \{\bu \in \bbR^2 \, : \, \alpha + g F(\bu) \geq 0\}$, while crucially the other two constraints are \emph{linear in $\nu$}.
All in all, plugging this in eq.~\eqref{eq:Sigma0_1} we have
\begin{align}
    \label{eq:Sigma0_2}
    &\tSigma_0(q, e) = \frac{-1 + (1 - 2\alpha) \log \alpha}{2} + \frac{1}{2} \log (1-q^2) \\ 
    \nonumber
    &+ \sup_{\substack{g > 0 \\ A > 0}} \sup_{\substack{\nu \, : \, \supp(\nu) \subseteq B_g \\ 
    \mcP(\nu)}}\left[- \log g - \frac{1}{2} \log A - g \EE_\nu[t(\bu)] 
    + \alpha \EE_{\nu} \log[\alpha + gF(\bu)] - \alpha H(\nu | \mu_q)\right],
\end{align}
where $\mcP(\nu)$ -- which also depends on all $(A, g, q, e)$ -- involves only \emph{linear} (equality and inequality) constraints over $\nu$.
We can thus introduce Lagrange multipliers for all these constraints.
We reach:
\begin{align}
    \label{eq:Sigma0_3}
    &\tSigma_0(q, e) = \frac{-1 + (1 - 2\alpha) \log \alpha}{2} + \frac{1}{2} \log (1-q^2) \\ 
    \nonumber
    &+ \sup_{\substack{g > 0 \\ A > 0}} \sup_{\nu \, : \, \supp(\nu) \subseteq B_g} \inf_{\substack{\lambda_A, \lambda_c, \lambda_e, \lambda_t \in \bbR \\ \lambda_h,\lambda_\star \geq 0}}\Bigg[- \log g - \frac{1}{2} \log A - g \EE_\nu[t(\bu)] 
    + \alpha \EE_{\nu} \log[\alpha + gF(\bu)] \\ 
    \nonumber
    &+ \alpha \lambda_A(A - \EE_\nu[A(\bu]) - \alpha \lambda_c \EE_\nu[c_q(\bu)] + \alpha \lambda_e (e - \EE_\nu[\ell(\bu)]) + \lambda_\star \EE_{\nu}[K_q(\bu)] 
    \\ 
    \nonumber
    & + \lambda_t \left(-\frac{1}{g} - \EE_\nu[t(\bu)]+ \alpha \EE_\nu \left[\frac{F(\bu)}{\alpha + g F(\bu)}\right] \right)
    \\ 
    \nonumber
    & - \lambda_h \left(\alpha \EE_{\nu} \left[\left(\frac{g F(\bu)}{\alpha + g F(\bu)}\right)^2\right]-1\right)
    - \alpha H(\nu | \mu_q)\Bigg].
\end{align}
For any fixed $(A, g)$, the function inside the variational problem in eq.~\eqref{eq:Sigma0_3} is: 
\begin{itemize}
    \item Strictly concave as a function of $\nu$ (it is the sum of a linear functional and of $-\alpha H(\nu|\mu_q)$, which is strictly concave). Moreover, notice that the constraint $\supp(\nu) \subseteq B_g$ is convex.
    \item Convex over $(\lambda_a, \lambda_c, \lambda_e, \lambda_t, \lambda_h)$, since it is linear. Moreover, the constraints $\lambda_h, \lambda_\star \geq 0$ are convex.
\end{itemize}
Therefore one can invert the corresponding supremum and infimum in eq.~\eqref{eq:Sigma0_3}.
As a last step, 
we use the Gibbs-Boltzmann variational formulation, which we recall here for a generic function $L$:
\begin{align}\label{eq:gibbs_principle}
    \begin{dcases}
    \argmax_{\nu \in \mcM_1^+(B)} \Big[\int \nu(\rd x) L(x) - \alpha H(\nu | \mu)\Big] 
    &= \frac{\mu(\rd x) \, e^{L(x)/\alpha}}{\int_B \mu(\rd x) \, e^{L(x)/\alpha}}, \\
    \sup_{\nu \in \mcM_1^+(B)} \Big[\int \nu(\rd x) L(x) - \alpha H(\nu | \mu)\Big] 
    &= \alpha \log \int_B \mu(\rd x) \, e^{L(x) /\alpha}.
    \end{dcases}
\end{align}
All in all, we reach eq.~\eqref{eq:Sigma_0_scalar}:
\begin{align}\label{eq:Sigma0_4}
    &\tSigma_0(q, e) 
    = \frac{-1 + (1-2\alpha) \log \alpha}{2} + \frac{1}{2} \log(1-q^2) \\ 
     \nonumber
     &+ \sup_{\substack{g > 0 \\ A > 0}} 
     \inf_{\substack{\lambda_A, \lambda_c, \lambda_e, \lambda_t\in \bbR \\ \lambda_h, \lambda_\star \geq 0}}
    \Bigg\{-\frac{1}{2} \log A + \alpha(\lambda_A A + \lambda_e e)
    - \log g - \frac{\lambda_t}{g} + \lambda_h \\
    &\nonumber
    + \alpha \log \int_{B_g} \mu_q(\rd \bu) \, e^{- \lambda_c c_q(\bu) - \lambda_A A(\bu) - \lambda_e \ell(\bu) + \log(\alpha + F(\bu) g)
    - \frac{g+\lambda_t}{\alpha} t(\bu) + \lambda_t \frac{F(\bu)}{\alpha + F(\bu) g}
    - \lambda_h\left(\frac{F(\bu) g}{\alpha + F(\bu) g}\right)^2 
    + \frac{\lambda_\star}{\alpha} K_q(\bu)
    } \Bigg\}.
\end{align}
We emphasize that the different scalar parameters appearing in eq.~\eqref{eq:Sigma0_4} have natural interpretations.
For instance, $g > 0$ represents the Stieltjes transform of the Hessian in $0$, i.e.\ the Stieltjes transform of $\mu_\alpha[\nu]$ at $z = t(\nu)$ (which real by hypothesis, since the Hessian spectral density is non-negatively supported). 
The Lagrange (KKT) multiplier $\lambda_h \geq 0$ enforces the constraint of eq.~\eqref{eq:g_smaller_than_edge}: as such, it ensures that $g$ is the solution to eq.~\eqref{eq:stieltjes_t_g} corresponding well to the Stieltjes transform of $\mu[\nu]$ evaluated at $t(\nu)$: indeed, as we discuss below (see the end of Section~\ref{subsubsec_app:var_to_scalar_TC}), this equation might have several solutions.
Finally, $\lambda_\star \geq 0$ enforces the additional constraint $\EE_{\nu}[K_q(\bu)] \geq 0$ that we impose in order to get a bound on the complexity of local minima which is tighter than the complexity of sub-extensive-index saddles.
The activation of this constraint is signaled by $\lambda_\star > 0$: when this is the case we have $\tSigma_0 < \Sigma_\sub$.

\subsubsection{Generic critical points}\label{subsubsec_app:var_to_scalar_TC}

We start from eq.~\eqref{eq:Sigma_TC}: 
\begin{align*}
    \Sigma_\TC(q, e) &= \frac{1 + \log \alpha}{2} + \frac{1}{2} \log (1-q^2) \\ 
    &\hspace{10pt}+ \sup_{\nu \in \mcM_\ell(q, e)}\left[-\frac{1}{2} \log \int_{\bbR^2} \nu(\rd \bu) A(\bu)
    + \kappa_{\alpha}(\nu) - \alpha H(\nu | \mu_q)\right].
\end{align*}
Similarly to the above derivation for local minima, we introduce Lagrange multipliers to fix the conditions in $\mcM_\ell(q,e)$.
We reach:
\begin{align}\label{eq:Sigma_TC_1}
    \nonumber
    \Sigma_\TC(q, e) &= \frac{1 + \log \alpha}{2} + \frac{1}{2} \log (1-q^2) + \sup_{A > 0} \sup_{\nu \in \mcP(\bbR^2)} \inf_{\lambda_A, \lambda_e, \lambda_c \in \bbR}\Bigg[-\frac{1}{2} \log A 
    + \alpha \lambda_A(A - \EE_\nu[A(\bu]) \\
    &- \alpha \lambda_c \EE_\nu[c_q(\bu)] + \alpha \lambda_e (e - \EE_\nu[\ell(\bu)])
    + \kappa_{\alpha}(\nu) - \alpha H(\nu | \mu_q)\Bigg].
\end{align}
We now tackle the term $\kappa_\alpha(\nu)$.
One has a similar formula to eq.~\eqref{eq:kappa_g_fixed}, see~\cite[Section~3.2]{maillard2020landscape}. Precisely, as $\eps \downarrow 0$, 
and with $g = g_r + i g_i$:
\begin{align}\label{eq:kappa_TC}
    \kappa_{\alpha}(\nu) = \extr_{g \in \bbC_+}\left[- \log |g| - t(\nu) g_r + \eps g_i + \alpha \EE_\nu \log |\alpha + F(\bu) g| - 1 - \alpha \log \alpha\right] + \smallO_\eps(1).
\end{align}
Notice that one can easily check that the extremization condition in eq.~\eqref{eq:kappa_TC} reduces to the Marchenko-Pastur equation, eq.~\eqref{eq:stieltjes_sample_covariance}: the corresponding $g \in \bbC_+$ is thus the Stieltjes transform of $\mu_\alpha[\nu]$ taken in $t(\nu) + i \eps$.
Following~\cite{maillard2020landscape}, we plug eq.~\eqref{eq:kappa_TC} in eq.~\eqref{eq:Sigma_TC_1}, 
and look for a saddle point of the resulting functional. Again, the saddle point equation on $\nu$ can be solved exactly, as we are extremizing a functional of the type of eq.~\eqref{eq:gibbs_principle}. We reach then eq.~\eqref{eq:Sigma_TC_scalar}:
\begin{align}
    \label{eq:Sigma_TC_2}
    \Sigma_\TC(q, e) &= \frac{-1 + (1- 2\alpha)\log \alpha}{2} + \frac{1}{2} \log(1-q^2) 
    + \extr_{\substack{A > 0, g \in \bbC_+ \\ 
    \lambda_A, \lambda_e, \lambda_c \in \bbR}}
    \Bigg[- \frac{1}{2} \log A - \log |g| + \eps g_i \\ 
    \nonumber
    &+ \alpha (\lambda_A A + \lambda_e e) 
    + \alpha \log \int_{\bbR^2} \mu_q(\rd \bu) e^{- \lambda_c c_q(\bu) - \lambda_A A(\bu) - \lambda_e \ell(\bu) + \log |\alpha + F(\bu) g| - \frac{g_r}{\alpha} t(\bu)}\Bigg].
\end{align}

\myskip 
\textbf{Remark: a difference with the previous formulas --}
Our formula for the complexity of all critical points (eq.~\eqref{eq:Sigma_TC_2}) is written as a global extremum, while the formula for local minima and sub-extensive saddles is rather written as a sup-inf involving more parameters, see eq.~\eqref{eq:Sigma0_4}.
The reason for this is eq.~\eqref{eq:kappa_TC}: indeed, while for any $\eps > 0$ eq.~\eqref{eq:kappa_TC} is known to have a single saddle point, on the other hand 
the equation
\begin{align*}
     - \frac{1}{g} - z g + \alpha \EE_\nu \left[\frac{F(\bu)}{\alpha + g F(\bu)}\right] = 0
\end{align*}
can admit more than one solution $g > 0$ if $z \notin \supp \mu_{\alpha}[\nu]$ (sometimes referred to as different ``branches'' of solutions to the Marchenko-Pastur equation, see e.g.~\cite{maillard2021large} or~\cite[Remark~2.6]{couillet2022random}).
Notice that this second branch is precisely ruled out by the condition of eq.~\eqref{eq:g_smaller_than_edge}.
The presence of this second branch motivated us to consider a nested max-min solver for the complexity of local minima (and sub-extensive index saddles), as we detail in Appendix~\ref{subsec_app:scalar_to_alg}. Since this problem is absent for the complexity of local minima, we rather consider there a simpler fixed-point solver for eq.~\eqref{eq:Sigma_TC_2}.
Nevertheless, we did encounter the presence of (at least) two fixed points to eq.~\eqref{eq:Sigma_TC_2} in a limited region of parameters, which we naturally associate to a phase transition in the complexity, see Section~\ref{subsec_app:phase_transition_total_complexity}, while we never observed such a phenomenon for minima and sub-extensive-index saddles. Investigating other forms of solvers for eq.~\eqref{eq:Sigma_TC_2}, including ones closer to the nested max-min solvers we use for local minima and sub-extensive-index saddles, is thus a very natural direction, which we plan to tackle in the future.

\subsection{From variational formulas to algorithms}\label{subsec_app:scalar_to_alg}

We show here how we compute in practice the solution to the variational principles for the complexities, eqs.~\eqref{eq:Sigma_0_scalar},\eqref{eq:Sigma_fin_scalar} and~\eqref{eq:Sigma_TC_scalar}.

\subsubsection{Local minima and saddles of sub-extensive index}

\textbf{The algorithm --}
We focus on $\tSigma_0(q, e)$ for concreteness, the case of saddle of sub-extensive index being very similar, and we comment on it afterwards.
Recall eq.~\eqref{eq:Sigma_0_scalar}:
\begin{align}\label{eq:Sigma_0_scalar_app}
    &\tSigma_0(q, e) 
    = \frac{-1 + (1-2\alpha) \log \alpha}{2} + \frac{1}{2} \log(1-q^2) \\ 
    \nonumber
     &+ \sup_{\substack{g > 0 \\ A > 0}} 
     \inf_{\substack{\lambda_A, \lambda_c, \lambda_e\in \bbR \\ \lambda_h, \lambda_t, \lambda_\star \geq 0}}
    \Bigg\{-\frac{1}{2} \log A + \alpha(\lambda_A A + \lambda_e e)
    - \log g - \frac{\lambda_t}{g} + \lambda_h \\
    \nonumber
    &
    + \alpha \log \int_{B_g} \mu_q(\rd \bu) \, e^{- \lambda_c c_q(\bu) - \lambda_A A(\bu) - \lambda_e \ell(\bu) + \log(\alpha + F(\bu) g)
    - \frac{g+\lambda_t}{\alpha} t(\bu) + \lambda_t \frac{F(\bu)}{\alpha + F(\bu) g}
    - \lambda_h\left(\frac{F(\bu) g}{\alpha + F(\bu) g}\right)^2 
    + \frac{\lambda_\star}{\alpha} K_q(\bu)
    } \Bigg\}.
\end{align}
As we emphasized in the derivation (and can be checked easily), the inner infimum over the Lagrange multipliers $\Lambda \coloneqq (\lambda_A, \lambda_c, \lambda_e, \lambda_h, \lambda_t, \lambda_\star)$ is convex, and can be solved efficiently for any value of $(A, g)$. 
In practice, we use convex minimization procedures (the limited memory-BFGS method~\cite{liu1989limited}) to solve it efficiently. 
We then apply another local maximization algorithm for the outer maximum in $(A, g)$, see Algorithm~\ref{alg:maxmin_minima}.
\begin{algorithm}[!t]
\SetAlgoLined
\emph{Initialize} variables $(A^\0, g^\0)$\;
\While{not converging}{
$\bullet$ \textbf{Step 1:} find $(\lambda_A^\topk, \lambda_c^\topk, \lambda_e^\topk, \lambda_t^\topk, \lambda_h^\topk, \lambda_\star^\topk)$ minimizing the convex functional
\begin{align*}
    &\alpha(\lambda_A A^\topk + \lambda_e e)
    \frac{\lambda_t}{g^\topk} + \lambda_h \\
    &
    + \alpha \log \int \mu_q(\rd \bu) \, e^{- \lambda_c c_q(\bu) - \lambda_A A(\bu) - \lambda_e \ell(\bu) + \log(\alpha + F(\bu) g^\topk)
    - \frac{g^\topk+\lambda_t}{\alpha} t(\bu) + \lambda_t \frac{F(\bu)}{\alpha + F(\bu) g^\topk}}
    \\ 
    & \hspace{100pt}
    \times e^{
    - \lambda_h\left(\frac{F(\bu) g^\topk}{\alpha + F(\bu) g^\topk}\right)^2 
    + \frac{\lambda_\star}{\alpha} K_q(\bu)
    }
\end{align*}
over $(\lambda_A, \lambda_c, \lambda_e, \lambda_t) \in \bbR$ and $(\lambda_h, \lambda^\star) \geq 0$.
\\
$\bullet$ \textbf{Step 2:} 
Make an update of $(A^\topk, g^\topk)$, using e.g.\ a projected gradient ascent algorithm or the L-BFGS algorithm. The 
gradient $(L_A, L_g)$ of the complexity functional in eq.~\eqref{eq:Sigma_0_scalar_app} is
\begin{align*}
    \begin{dcases}
    L_A &= - \frac{1}{2A} + \alpha \lambda_A, \\
       L_g &= -\frac{1}{g} + \frac{\lambda_t^\topk}{g^2} 
       -\EE_{\hat{\nu}^\topk}[t(\bu)] + \alpha \EE_{\hat{\nu}^\topk}\left[\frac{F(\bu)}{\alpha + g F(\bu)} \right] 
        \\ 
        \nonumber
        &
        - \alpha \lambda_t^\topk \EE_{\hat{\nu}^\topk}\left[\frac{F(\bu)^2}{(\alpha + g F(\bu))^2} \right]
        - 2 \alpha^2 g \lambda_h^\topk \EE_{\hat{\nu}^\topk}\left[\frac{F(\bu)^2}{(\alpha + g F(\bu))^3} \right].
    \end{dcases}
\end{align*}
Here $\hnu^\topk$ is the estimate of the joint law of the labels (eq.~\eqref{eq:nu0}) with the current estimate of the Lagrange multipliers and the given values of $(A, g)$. \;
$k = k + 1$\;
}
\caption{An algorithm to compute $\tSigma_0(q, e)$.
\label{alg:maxmin_minima}
}
\end{algorithm}

\myskip 
\textbf{Gradients --}
Notice that in Step $1$ of the algorithm, the gradient $G_\theta$ of the functional with respect to the Lagrange multipliers $\lambda_\theta$ (for $\theta \in \{A, c, e, t, h, \star\}$) is simply: 
\begin{align}
    \begin{dcases}
        G_A &= \alpha[a - \EE_{\hnu}[A(\bu)]], \\
        G_c &= - \alpha\EE_{\hnu}[c_q(\bu)], \\
        G_e &= \alpha[e - \EE_{\hnu}[\ell(\bu)]], \\
        G_t &= - \frac{1}{g} - \EE_{\hnu}[t(\bu)] + \alpha\EE_{\hnu}\left[\frac{F(\bu)}{\alpha + F(\bu) g}\right] , \\
        G_h &= 1 - \alpha\EE_{\hnu}\left[\left(\frac{F(\bu) g}{\alpha + F(\bu) g}\right)^2\right], \\
        G_\star &= \EE_{\hnu}[K_q(\bu)].
    \end{dcases}
\end{align}
%
The gradients with respect to $(A, g)$ are given in Algorithm~\ref{alg:maxmin_minima}.
Notice that that the condition imposed by $\lambda_h$ (eq.~\eqref{eq:g_smaller_than_edge}) implies that if $\bu \sim \nu$, then $X \coloneqq \alpha + g F(\bu) \geq 0$ either: $(i)$ is not supported around $0$, or $(ii)$ has a quickly decaying density around $0$ such that $\EE[F^2 / (\alpha + g F)^2] \leq 1$.
This implies that when differentiating the complexity functional in eq.~\eqref{eq:Sigma_0_scalar_app} with respect to $g$ there is no contribution of the boundary term 
$\alpha \EE_\nu[F(\bu) \delta(\alpha + g F(\bu))]$.

\myskip
In practice, we use as well the L-BFGS algorithm for the outer maximization, re-parametrizing the functional in terms of $\log g$ and $\log A$ to ensure positivity of $A, g$. While such a change of variable may affect concavity properties, we never encountered the presence of different local maxima (by varying initial conditions), and it allows to impose the positivity constraint in a straightforward manner.
We emphasize, however, that these empirical observations do not constitute a formal convergence guarantee: nevertheless, within the scope of our experiments, the method proved sufficiently stable and efficient to explore the landscape of solutions of interest.

\myskip
\textbf{Inner integrals and computation time --}
The inner integrals appearing in the objective function are evaluated numerically using SciPy’s adaptive quadrature routines. Since these two-dimensional integrals must be recomputed repeatedly during the optimization process, their evaluation is accelerated using Numba’s just-in-time (JIT) compilation~\cite{lam2015numba}. In particular, the integrand functions are compiled to optimized machine code prior to the numerical integration, significantly reducing the per-evaluation computational cost.

\myskip
These numerically evaluated integrals are embedded within the outer maximization procedure used to solve eq.~\eqref{eq:Sigma_0_scalar_app}. Owing to the efficiency gains provided by JIT compilation and adaptive quadrature, the outer maximization converges rapidly, typically requiring on the order of ten iterations. Overall, the full solution of eq.~\eqref{eq:Sigma_0_scalar_app} is typically obtained in around one minute on a standard laptop-class CPU.

\myskip 
\textbf{Saddles of sub-extensive index --}
The computation of the complexity $\Sigma_\fin(q, e)$ of sub-extensive-index saddles (eq.~\eqref{eq:Sigma_fin_scalar}) is extremely similar to the above: one simply imposes $\lambda_\star = 0$ in Algorithm~\ref{alg:maxmin_minima}.

\subsubsection{Critical points}

On the other hand, for the complexity of all critical points $\Sigma_\TC(q, e)$ (eq.~\eqref{eq:Sigma_TC_scalar}), we simply use an iterative procedure, 
as we discussed in Section~\ref{subsubsec_app:var_to_scalar_TC}. Each step of the procedure proceeds as follows. 
We denote $\hnu^\topk$ the measure $\nu_\tot$ in eq.~\eqref{eq:nu_tot} with the current estimate of the variables at iteration $k$.
\begin{enumerate}[label=$( \roman* )$]
    \item Update 
    \begin{equation*}
        g^\topkpone = - \left[\EE_{\hnu^\topk}[t(\bu)] + i \eps - \alpha \EE_{\hnu^\topk}\left(\frac{F(\bu)}{\alpha + g^\topk F(\bu)}\right)  \right]^{-1}.
    \end{equation*}
    \item Update $A^\topkpone = \EE_{\hnu^\topk}[A(\bu)]$.
    \item Find $(\lambda_c^\topkpone, \lambda_e^\topkpone)$ satisfying the constraints 
    \begin{equation*}
        \begin{dcases}
            \EE_{\hnu[\lambda_c^\topkpone, \lambda_e^\topkpone]}(c_q(\bu)) &= 0, \\
            \EE_{\hnu[\lambda_c^\topkpone, \lambda_e^\topkpone]}(\ell(\bu)) &= e.
        \end{dcases}
    \end{equation*}
    \item Update $\lambda_A^\topkpone = 1 / (2 \alpha A^{\topkpone})$.
\end{enumerate}
These equations correspond to iterations aiming to find a zero of the gradient of the functional with respect to (respectively) $g, \lambda_a, \lambda_c, \lambda_e, A$, and are similar to the ones stated in~\cite{maillard2020landscape} in a simpler model. Notice that in the penultimate step we actually require $(\lambda_c^\topk, \lambda_e^\topk)$ to be exact solutions to the constraints (which may require several iterations). As we saw, this is a convex optimization problem and does not cause any numerical issues.  
Alternatively to $(iv)$, one can also update $\lambda_A$ together with these other parameters (causing a benign $3$-dimensional convex optimization problem): we did not find it to affect convergence properties, but increased slightly the time per iteration.

\myskip 
Although this iterative scheme is not derived from a local optimization principle (since the extremum in eq.~\eqref{eq:Sigma_TC_scalar} is not in the form of a $\sup\inf$), and thus remains somewhat heuristic,
we found it to possess good convergence properties. Using similar procedures to the computation of the two-dimensional integrals, we found the iterations to typically converge in a few $100s$ to a few $1000s$ iterations, which can take from a few seconds to a few minutes on a standard laptop-class CPU, depending on the parameters of the problem and the required stopping tolerance.

\subsubsection{Further details and remarks}

We give here a few other remarks regarding the implementation of Algorithm~\ref{alg:maxmin_minima} (and its counterpart for sub-extensive-index saddles).

\begin{itemize}[leftmargin=*]
    \item Consider the gradient $L_g$ of the complexity functional given in Algorithm~\ref{alg:maxmin_minima}.
    At a global maximizer $(A, g)$, we have $L_g = 0$. However, the constraint of eq.~\eqref{eq:stieltjes_t_g} imposes that then: 
    \begin{align}\label{eq:Lg_0}
        \lambda_t \left(\frac{1}{g^2} 
        - \alpha \EE_{\hat{\nu}}\left[\frac{F(\bu)^2}{(\alpha + g F(\bu))^2} \right]\right)
        - 2 \alpha^2 g \lambda_h \EE_{\hat{\nu}}\left[\frac{F(\bu)^2}{(\alpha + g F(\bu))^3} \right] = 0.
    \end{align}
    One can separate two cases:
    \begin{enumerate}[label=$( \roman* )$, leftmargin=*]
        \item Either the constraint of eq.~\eqref{eq:g_smaller_than_edge} is not saturated (i.e.\ the left-hand side is strictly smaller than $1$): then the KKT multiplier $\lambda_h = 0$. By eq.~\eqref{eq:Lg_0} we reach then that $\lambda_t = 0$.
        In this setting the positivity constraint on the Hessian's density in the supremum over the law $\nu$ of the labels is not saturated at the maximum.
        \item Either the constraint of eq.~\eqref{eq:g_smaller_than_edge} is saturated, but then eq.~\eqref{eq:Lg_0} implies $\lambda_h = 0$.
    \end{enumerate}
    We therefore always have $\lambda_h = 0$ at the optimal value of the variational principle. Notice that the presence of the KKT multiplier $\lambda_h \geq 0$ during the optimization process is still very important: as we saw, it allows e.g.\ to remove spurious local maxima associated to the second branch of solutions to the Marchenko-Pastur equation.
    \item According to the remark above, at a global maximizer $(A, g)$, if there exists $\bu \in \bbR^2$ such that $\alpha + g F(\bu) < 0$, then we must have $\lambda_t \geq 0$ at this value of $(A, g)$, in order for the density of $X = \alpha + g F(\bu)$ (for $\bu \sim \nu$) to decay as $X \downarrow 0$.
    Similarly, at a maximizer, we also have $L_A = 0$, and thus $\lambda_A = (2\alpha A)^{-1} > 0$.
    In practice, we impose $\lambda_A \geq 0$ when computing Step~1 in Algorithm~\ref{alg:maxmin_minima}, and $\lambda_t \geq 0$ when there exists $\bu$ with $\alpha + g F(\bu) < 0$, as we found it to greatly improve numerical stability.
    Moreover, by convexity, these additional constraints do not affect the solution to the variational problem as long as they are not saturated at the optimum 
    (i.e.\ $\lambda_A > 0$, $\lambda_t > 0$), which we always found to hold.
\end{itemize}

%% file: sections/appendix/bbp.tex
We derive here the condition for the ``BBP transition'' in the Hessian's spectrum, i.e.\ the appearance of an isolated eigenvalue at the left of the density's bulk, and whose eigenvector is positively correlated to the signal $\btheta^\star$. 
We assume to have a given law $\nu(y, y^\star)$, corresponding to the limiting empirical law of the labels on the type of points we are considering (e.g.\ saddles of finite index, or critical points at a given loss value).
We also assume we are given an
overlap value $q \in [0, 1)$: in this regard, this computation generalizes the derivation of~\cite{bonnaire2025role} that holds at $q = 0$. 

\myskip
Let us recall some notations: $q = \btheta \cdot \btheta^\star$
and $\|\btheta\| = \|\btheta^\star\| = 1$. The data vectors are $\bx_1, \cdots, \bx_n \iid \mcN(0, \Id_d)$, 
and we denote $y_i^\star \coloneqq \bx_i \cdot \btheta^\star$ and $y_i \coloneqq \bx_i \cdot \btheta$.
Recall $n/d \to \alpha > 1$.
Up to an additive shift (which is irrelevant in terms of transition for an isolated eigenvalue), the Hessian is (see eq.~\eqref{eq:grad_Hess}): 
\begin{align}\label{eq:Hessian_form}
    \bH = \frac{1}{n} \sum_{i=1}^n \partial_1^2 \ell(y_i, y_i^\star) (P_\btheta^\perp \bx_i) (P_\btheta^\perp \bx_i)^\T,
\end{align}
where $P_\btheta^\perp$ is the orthogonal projection on $\{\btheta\}^\perp$. 
Without loss of generality, by rotational invariance of the Gaussian distribution we can assume that 
\begin{align}
    \begin{dcases}
        \btheta^\star &= (1, 0, \cdots, 0), \\
        \btheta &= (q, \sqrt{1-q^2}, \cdots, 0).
    \end{dcases}
\end{align}
Notice that we can decompose $\bx_i$ as 
\begin{align}
    \bx_i = \left(y_i^\star, \frac{y_i - q y_i^\star}{\sqrt{1-q^2}}, 0, \cdots, 0\right) + \bv_i,
\end{align}
where $\bv_i$ is a Gaussian standard vector in $\{\btheta^\star, \btheta\}^\perp$, independent of $(y_i, y_i^\star)$.
Moreover, 
\begin{align}\label{eq:def_bzi}
   \bz_i \coloneqq P_\btheta^\perp \bx_i = \frac{y_i^\star - q y_i}{\sqrt{1-q^2}} \tbw + \bv_i,
\end{align}
where 
\begin{equation}\label{eq:tbw}
\tbw \coloneqq \frac{P_\btheta^\perp (\btheta^\star)}{\|P_\btheta^\perp (\btheta^\star)\|} = \left(\sqrt{1-q^2},-q, 0, \cdots, 0\right) \in \{\btheta\}^\perp.
\end{equation}
Let $\bG(z) \coloneqq (\bH - z \Id_d)^{-1}$ be the resolvent of $\bH$, and $g_d(z) \coloneqq (1/d) \Tr \bG(z)$ its Stieltjes transform.
Using the relation $(\bH - z \Id_d) \bG = \Id_d$, we get the equation 
\begin{align}\label{eq:resolvent}
    - z g_d(z) + \frac{1}{n} \sum_{i=1}^n \partial_1^2 \ell(y_i, y^\star_i) \frac{\bz_i^\T \bG \bz_i}{d} = 1.
\end{align}

\myskip 
\textbf{The bulk --}
The computation of the bulk of the Hessian is classical~\cite{marchenko1967distribution,silverstein1995empirical}, and can be obtained by the cavity method from eq.~\eqref{eq:resolvent}). 
One finds that the limiting Stieltjes transform $g(z) = \lim_{d \to \infty} g(z)$ of the limiting spectral density of $\bH$ (denoted $\sigma(x)$) satisfies, for any $z \in \bbC_+$, and assuming the empirical law of $(y_i, y^\star_i)$ converges to $\nu$: 
\begin{align}\label{eq:general_marchenko_pastur}
    z = - \frac{1}{g(z)} + \alpha \EE_{(y, y^\star)\sim \nu} \left[\frac{f(y, y^\star)}{\alpha + g(z) f(y, y^\star)}\right],
\end{align}
with $f(y, y^\star) \coloneqq \partial_1^2 \ell(y, y^\star)$.
Eq.~\eqref{eq:general_marchenko_pastur} is sometimes called the Marchenko-Pastur equation.
Recall that the density $\sigma(x)$ is directly related to the asymptotic spectral density $\rho(w)$ of the spherical Hessian of the empirical loss, with a simple additive shift: 
\begin{equation*}
    \sigma(x) = \rho(x - t(\nu)),
\end{equation*}
with $t(\nu)$ defined in eq.~\eqref{eq:def_kappa_t}.

\myskip
The left edge of the bulk $x_{\min}$ is also obtained in a classical fashion, see Appendix~\ref{subsec_app:Hessian_spectrum}. 
Essentially, under mild conditions on the behavior of the law of $f(y, y^\star)$ near its minimal value (for $(y, y^\star) \sim \nu)$,
the Stieltjes transform $g_{\min} = g(x_{\min})$ of the left edge of the bulk is 
the positive solution to the equation 
\begin{align}\label{eq:gmin}
    1 &= \alpha \, \EE_{y, y^\star}  \left[\left(\frac{g_{\min} f(y, y^\star)}{\alpha + g_{\min} f(y, y^\star)}\right)^2\right].
\end{align}
One can then obtain the bottom edge $x_{\min}$ by plugging $g_{\min} = g(x_{\min})$ in eq.~\eqref{eq:general_marchenko_pastur}.

\myskip 
\textbf{The outlier --}
We follow a close computation to the one of~\cite{bonnaire2025role}. Notice that the Hessian lives in the tangent space to $\btheta$: therefore, 
an outlier correlated with $\btheta^\star$ actually means an outlier in the direction $\tbw$ defined in eq.~\eqref{eq:tbw}. We denote $\tg \coloneqq \tbw^\T \bG(z) \tbw$.
In a similar fashion to eq.~\eqref{eq:resolvent}, we get
\begin{align}\label{eq:cavity_outlier}
    -z \tg + \frac{1}{n} \sum_{i=1}^n f(y_i, y^\star_i) (\tbw^\T \bz_i) \cdot (\bz_i^\T \bG \tbw) = 1.
\end{align}
From eq.~\eqref{eq:def_bzi}, we have $ \tbw^\T \bz_i = (y_i^\star - q y_i) / \sqrt{1-q^2}$. Further, denoting 
\begin{align}\label{eq:def_bg_mini}
    \bG_{- i} \coloneqq \left( \frac{1}{n} \sum_{j (\neq i)} f(y_j, y_j^\star) \bz_j \bz_j^\T- z \Id_d\right)^{-1},
\end{align}
we get using the Shermann-Morrison formula: 
\begin{align}\label{eq:shermann_morrison}
    \bz_i^\T \bG \tbw = \frac{\bz_i^\T \bG_{-i} \tbw}{1 + \frac{1}{n} f(y_i, y_i^\star) \bz_i^\T \bG_{-i} \bz_i}.
\end{align}
We have $\bz_i^\T \bG_{-i} \bz_i \simeq d g(z)$ at leading order as $d \to \infty$ (since $\bv_i$ and $\bG_{-i}$ are independent, and the other term of $\bz_i$ in eq.~\eqref{eq:def_bzi} only contributes to sub-leading order).
Moreover, again as $d \to \infty$:
\begin{align}\label{eq:zi_G_w}
    \bz_i^\T \bG_{-i} \tbw = \frac{y_i^\star - q y_i}{\sqrt{1-q^2}} \underbrace{\tbw^\T \bG_{-i} \tbw}_{\simeq \tg} + \bv_i^\T \bG_{-i} \tbw.
\end{align}
Plugging eqs.~\eqref{eq:shermann_morrison} and \eqref{eq:zi_G_w} in eq.~\eqref{eq:cavity_outlier}, we get
\begin{align}\label{eq:cavity_outlier_2}
    -z \tg + \frac{1}{n} \sum_{i=1}^n \frac{f(y_i, y^\star_i)}{1 + \frac{g(z)}{\alpha} f(y_i, y^\star_i)} \frac{(y_i^\star - qy_i)}{\sqrt{1-q^2}} \left[ \frac{y_i^\star - q y_i}{\sqrt{1-q^2}}\tg + \bv_i^\T \bG_{-i} \tbw\right] = 1 + \smallO_d(1).
\end{align}
Notice that $\bv_i$ is a random Gaussian vector, independent of $(y_i, y_i^\star)$ and of $\bG_{-i}$. Taking expectation in eq.~\eqref{eq:cavity_outlier_2} with respect to $\bv_i$ 
(again, we assume concentration 
of $g, \tg$ as $d \to \infty$)
yields that the last term in eq.~\eqref{eq:cavity_outlier_2}
only contributes to sub-leading-order as $d \to \infty$. 
%

\myskip
In the end
we get, taking $d \to \infty$: 
\begin{align}\label{eq:cavity_outlier_3}
    - \frac{1}{\tg} = z - \frac{\alpha}{1-q^2} \EE_{(y, y^\star) \sim \nu} \left[\frac{(y^\star - q y)^2 f(y, y^\star)}{\alpha + g(z) f(y, y^\star)}\right].
\end{align}
The presence of an outlier correlated with $\tbw$ is signaled by the singularity of the resolvent, i.e.\ the divergence of $|\tg|$ (as argued in~\cite{bonnaire2025role}).
Using this criterion in eq.~\eqref{eq:cavity_outlier_3}, we reach that the equation satisfied by an outlier eigenvalue $x^\star$ is 
\begin{equation}\label{eq:outlier}
    x^\star = \frac{\alpha}{1-q^2} \EE_{(y, y^\star) \sim \nu} \left[\frac{(y^\star - q y)^2 f(y, y^\star)}{\alpha + g(x^\star) f(y, y^\star)}\right],
\end{equation}
where $g(x^\star)$, the limiting density's Stieltjes transform, can be computed by solving eq.~\eqref{eq:general_marchenko_pastur}.

\myskip 
\textbf{Conclusion --}
For a given estimate of the asymptotic empirical law of $(y_i, y^\star_i)_{i=1}^n$, and given values of $(\alpha, q)$, the threshold $\alpha_{\BBP}$ is given by the smallest value of $\alpha$ such that $x^\star = x_{\min}$.
In particular, at this point we have, with $g_{\min}$ the solution of eq.~\eqref{eq:gmin}, that $x_{\min} = x_2$, with
\begin{align}\label{eq:BBP_condition}
    \begin{dcases}
        x_{\min} &\coloneqq - \frac{1}{g_{\min}}  + \alpha \EE_{y, y^\star} \left[\frac{f(y, y^\star)}{\alpha + g_{\min} f(y, y^\star)}\right], \\ 
        x_2 &\coloneqq \frac{\alpha}{1-q^2} \EE_{y, y^\star} \left[\frac{(y^\star - q y)^2 f(y, y^\star)}{\alpha + g_{\min} f(y, y^\star)}\right].
    \end{dcases}
\end{align}
The equality of the two terms in eq.~\eqref{eq:BBP_condition} is what we numerically evaluate to probe the BBP transition, see Section~\ref{subsec_app:bbp_kr}.
Notice that generically $x_2 \neq x^\star$ (because of the different Stieltjes transform used in the denominator):
they only coincide at the BBP transition point.

%% file: sections/appendix/more_kac_rice.tex
In this appendix, we present a deeper exploration of the landscape of phase retrieval using the Kac-Rice formalism, of which we showed some essential results in Section~\ref{sec:kac_rice}: 
\begin{itemize}
    \item In Section~\ref{subsec_app:phase_diagram_general_a}, we show the generalization of the topological phase diagram (Fig.~\ref{fig:phase_diagram}) to values of $a \in \{0.1, 1.0\}$, where the same phenomenology remains.
    \item In Section~\ref{subsec_app:band_high_overlap}, we uncover loss values below which all local minima 
    are located in a band with positive correlation with the signal $\btheta^\star$, a phenomenon previously observed in Gaussian landscapes~\cite{Ros2019}.
    \item In Section~\ref{subsec_app:phase_transition_total_complexity}, we describe a first-order phase transition observed for the complexity of \emph{all} critical points. 
    \item Finally, in Section~\ref{subsec_app:bbp_kr}, we provide examples of the computation of the BBP transition in the ``BBP-KR'' formalism.
\end{itemize}



\subsection{The phase diagram for other values of $a$}\label{subsec_app:phase_diagram_general_a}

\begin{figure}[!t]
  \begin{subfigure}{\linewidth}
    \centering
    \includegraphics[width=\linewidth]{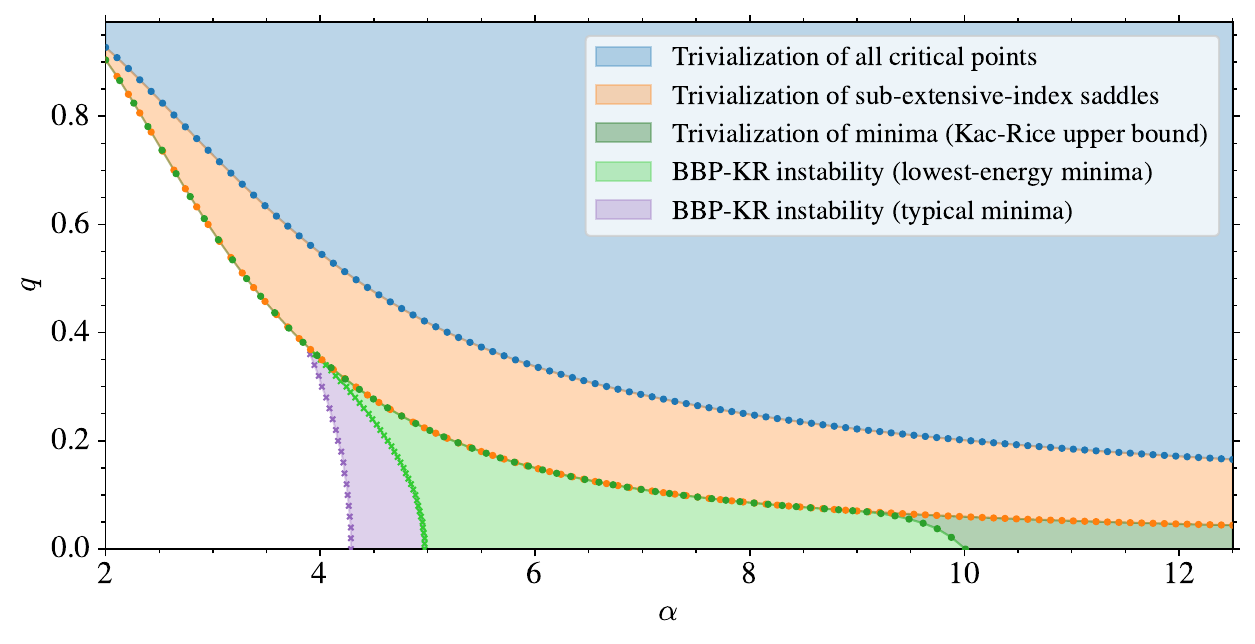}
   \caption{
    The phase diagram for $a = 0.1$.
   }
   \label{subfig:phase_diagram_a_0.1}
  \end{subfigure}

  \medskip 

  \begin{subfigure}{\linewidth}
    \centering
    \includegraphics[width=\linewidth]{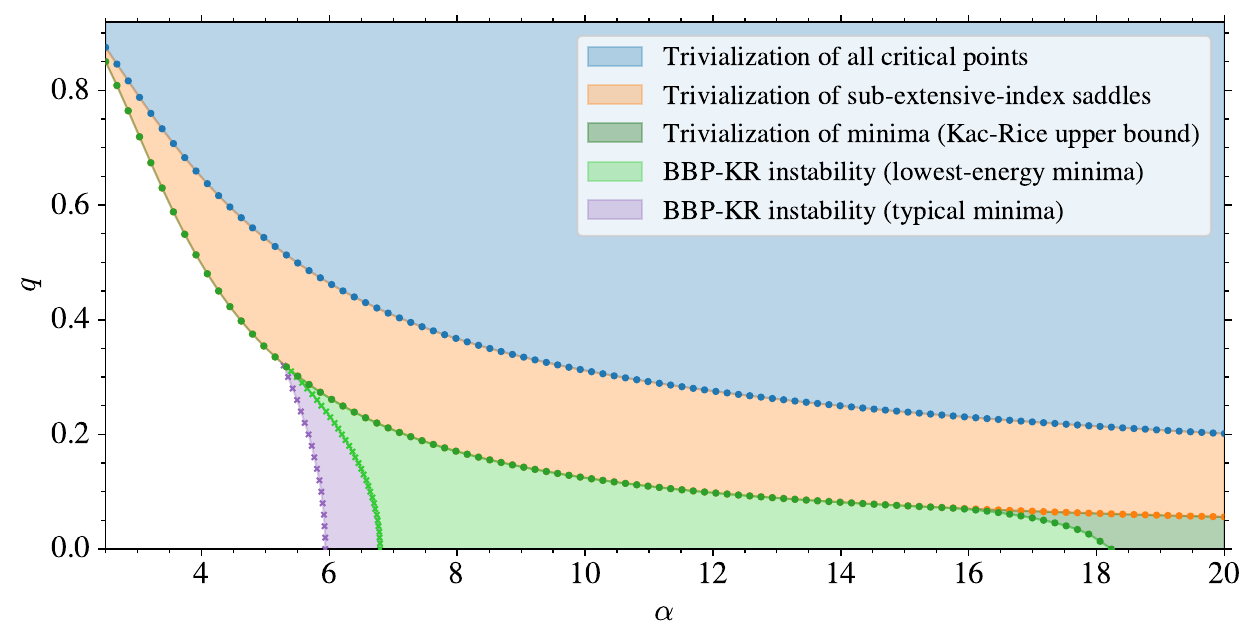}
   \caption{
    The phase diagram for $a = 1.0$.
   }
   \label{subfig:phase_diagram_a_1.0}
  \end{subfigure}
    \caption{
        Phase diagram predicted by the Kac-Rice method, for $a \in \{0.1, 1.0\}$. The conventions and colors are the same as in Fig.~\ref{fig:phase_diagram} in the main text.
        \label{fig:phase_diagram_various_a}
    }
\end{figure}

In Fig.~\ref{fig:phase_diagram_various_a} we generalize the topological phase diagram of Fig.~\ref{fig:phase_diagram}
to other values of $a \in \{0.1, 1.0\}$. The same phenomenology is retained: we simply observe that the sample complexity thresholds for trivialization of different types of critical points, and for the BBP transitions, are generically larger as $a$ increases.

\subsection{Appearance of local minima at high overlap}\label{subsec_app:band_high_overlap}

We did not find evidence in the annealed complexity for the emergence of minima at high overlap with the signal when there is no local minima at $q = 0$: we always found the complexities $\Sigma(q)$ to decrease monotonically with $q$, contrary e.g.\ to what happens in some Gaussian models of random landscapes~\cite{Ros2019}.
On the other hand, we did exhibit examples where the complexity $\Sigma(q, e)$ has a non-mononotic behavior with $q$, if the loss (or energy) value $e$ is small enough.
We illustrate it in Fig.~\ref{fig_app:band_high_overlap}:
\begin{figure}[!h]
    \centering
    \includegraphics[width=0.75\textwidth]{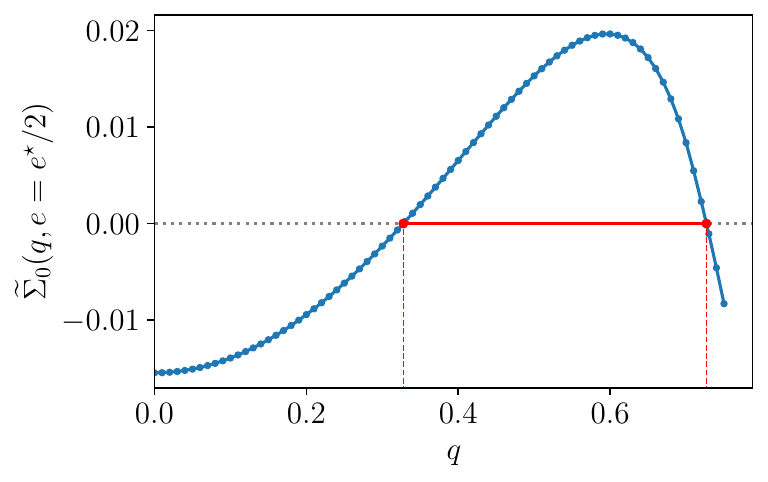}
    \caption{
        \centering
        For $a = 1.0$ and $\alpha = 3.0$, the annealed complexity $\tSigma_0(q, e)$ as a function of $q$, 
        for $e = e_\star(q = 0) / 2 \simeq 0.048$, half of the loss value of typical minima. In red we show the overlap band of positive complexity.
        \label{fig_app:band_high_overlap}
    }
\end{figure}
we consider $a = 1.0$ and $\alpha = 3.0$. At this value the complexity $\tSigma_0(q = 0) > 0$, see Fig.~\ref{subfig:phase_diagram_a_1.0}, and the Kac-Rice formula predicts that local minima have a typical energy $e^\star(q = 0)$. We then fixed $e = e^\star(q = 0) / 2$, and show the function $q \mapsto \Sigma(q, e)$. 
Interestingly, the annealed complexity predicts that for these low loss values (i.e.\ further down in the landscape), local minima are located at high overlap,
a phenomenon which is not visible when counting all minima: in this case, the most numerous are always located at the equator (i.e.\ at $q = 0$).

\subsection{A phase transition for the complexity of all critical points}\label{subsec_app:phase_transition_total_complexity}

\begin{figure}[!t]
    \centering
    \includegraphics[width=1.0\textwidth]{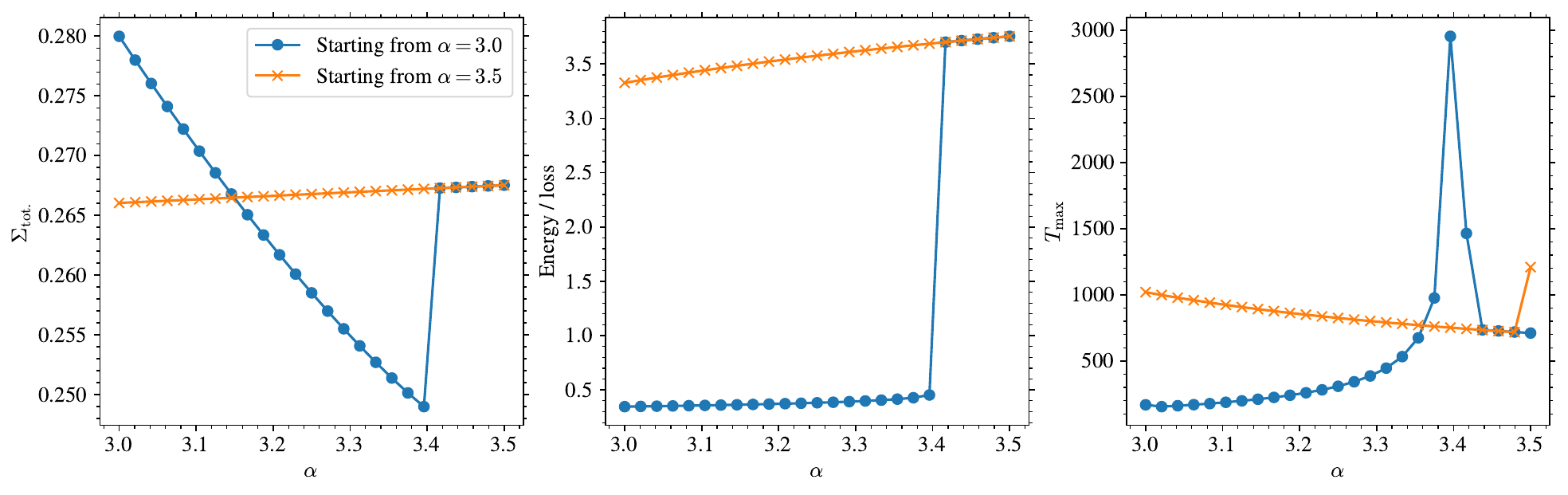}
    \caption{
        \centering
        Evolution of (left) the total complexity $\Sigma_\TC$, (middle) the energy, and (right) total number of steps as functions of $\alpha$ for two starting points: $\alpha\in\{3.0, 3.5\}$. 
        \label{fig:phase_transition_TC}
    }
\end{figure}

In Fig.~\ref{fig:phase_transition_TC}, we display the evolution of the total complexity $\Sigma_\TC$, the energy and the total number of steps as functions of $\alpha$ for two starting points to solve the equations numerically, either $\alpha=3.0$ or $\alpha=3.5$. It clearly shows that the iterative algorithm for critical points (eq.~\eqref{eq:Sigma_TC_scalar}) might have more than one fixed point, as we briefly pointed out in Section~\ref{subsubsec:final_remarks}. Indeed, starting from $\alpha=3.0$, the algorithm stays on a low-energy branch with decreasing complexity until $\alpha\approx3.4$ where it jumps to the high-energy branch, associated with a critical slowing down shown in the total number of steps on the right. This suggest the existence of a first-order phase transition for the complexity of all critical points.
However, in our investigations we did not find any numerical evidence of the presence of multiple local maxima in $(g, a)$ of eqs.~\eqref{eq:Sigma_0_scalar} and~\eqref{eq:Sigma_fin_scalar}, our formulas for the complexity of local minima and sub-extensive-index saddles.

\subsection{The BBP-KR instability}\label{subsec_app:bbp_kr}

\begin{figure}
    \centering
    \includegraphics[width=1.0\textwidth]{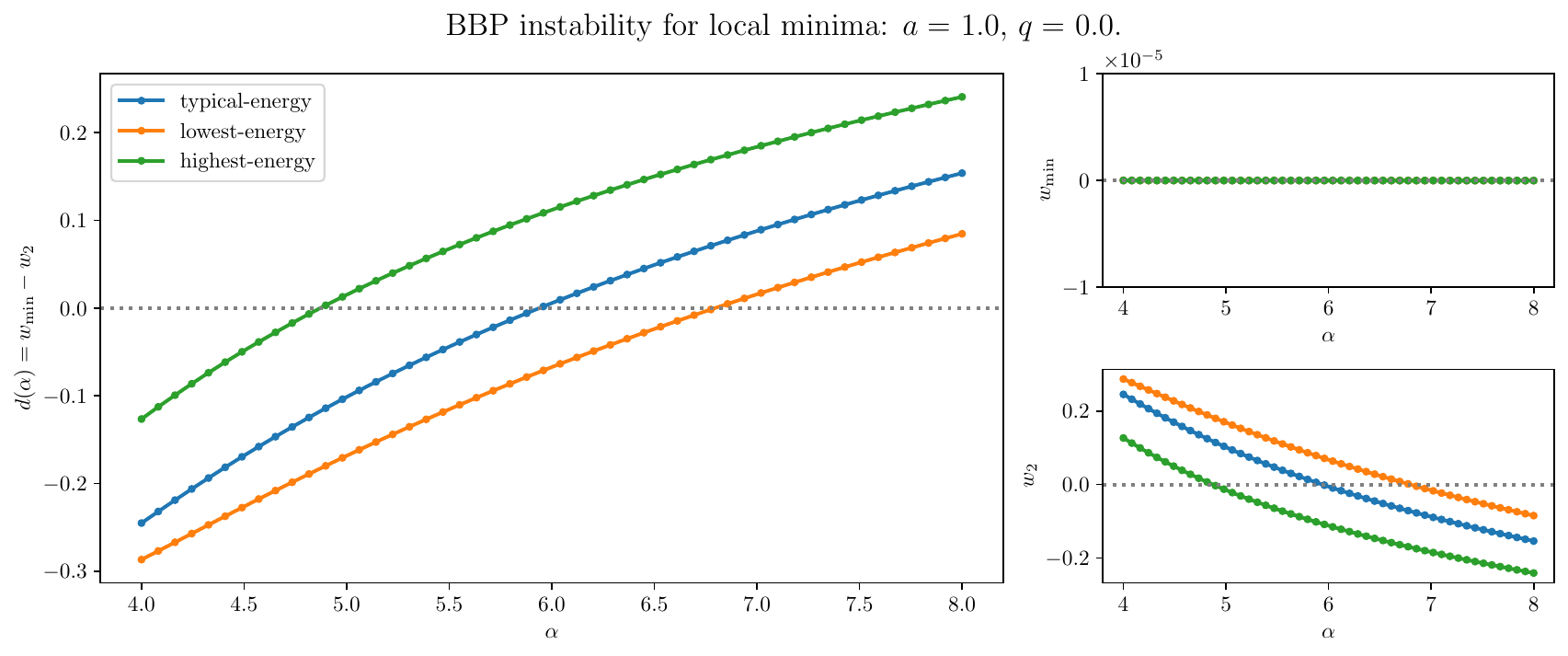}
    \caption{
        \centering
        For $a = 1.0$ and $q = 0.0$, we show the computation of the BBP transition point for typical-energy, lowest-energy and highest-energy minima.
        \label{fig:BBP_computation_example}
    }
\end{figure}
In Fig.~\ref{fig:BBP_computation_example} we show an example of a computation of the BBP transition, whose theory is given in Section~\ref{sec_app:bbp}.
We plot $d(\alpha) \coloneqq w_{\min} - w_2 = x_{\min} - x_2$: we see it crosses zero at a finite $\alpha$, which is larger as we consider larger-energy minima.
Recall that $w_{\min} = x_{\min} - t(\nu)$ is really the left edge of the Hessian's asymptotic spectral density.
We emphasize in particular the top-right plot in Fig.~\ref{fig:BBP_computation_example}: it shows that despite the appearance of the BBP instability, 
the bulk of the Hessian touches $0$ for all values of $\alpha$ considered.


%% file: sections/appendix/dynamics_general_a.tex
\begin{figure}
    \centering
    \includegraphics[width=.9\textwidth]{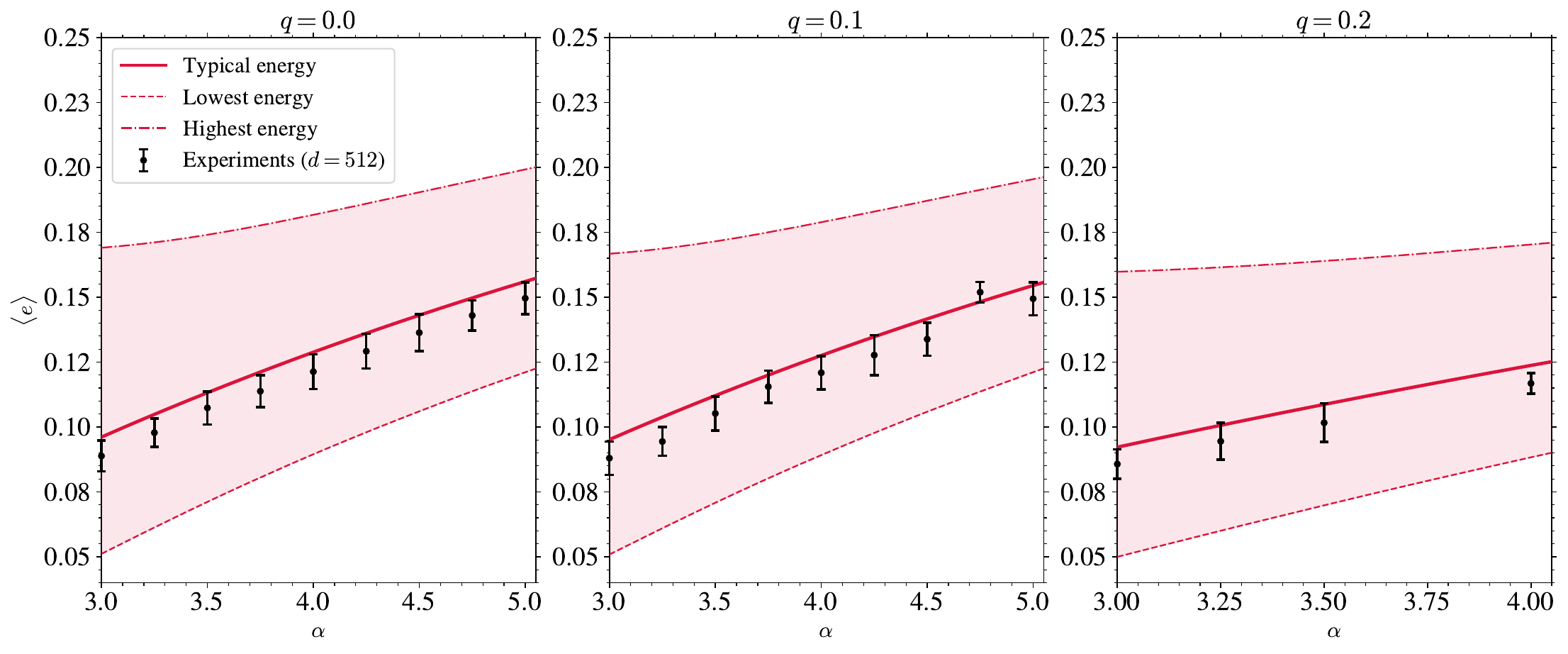}
    \caption{
        \centering Evolution of the average energy $\langle e \rangle$ of the minima for $a=1$ with $\alpha$ at fixed (left) $q=0.0$, (middle) $q=0.1$, and (right) $q=0.2$ obtained from the experiments (in black) and the band of energy predicted by the annealed Kac-Rice (in shaded red).
        \label{fig:energies_a1.0_KR_exp}
    }
\end{figure}
\begin{figure}
  \begin{subfigure}{\linewidth}
    \centering
    \includegraphics[width=.9\linewidth]{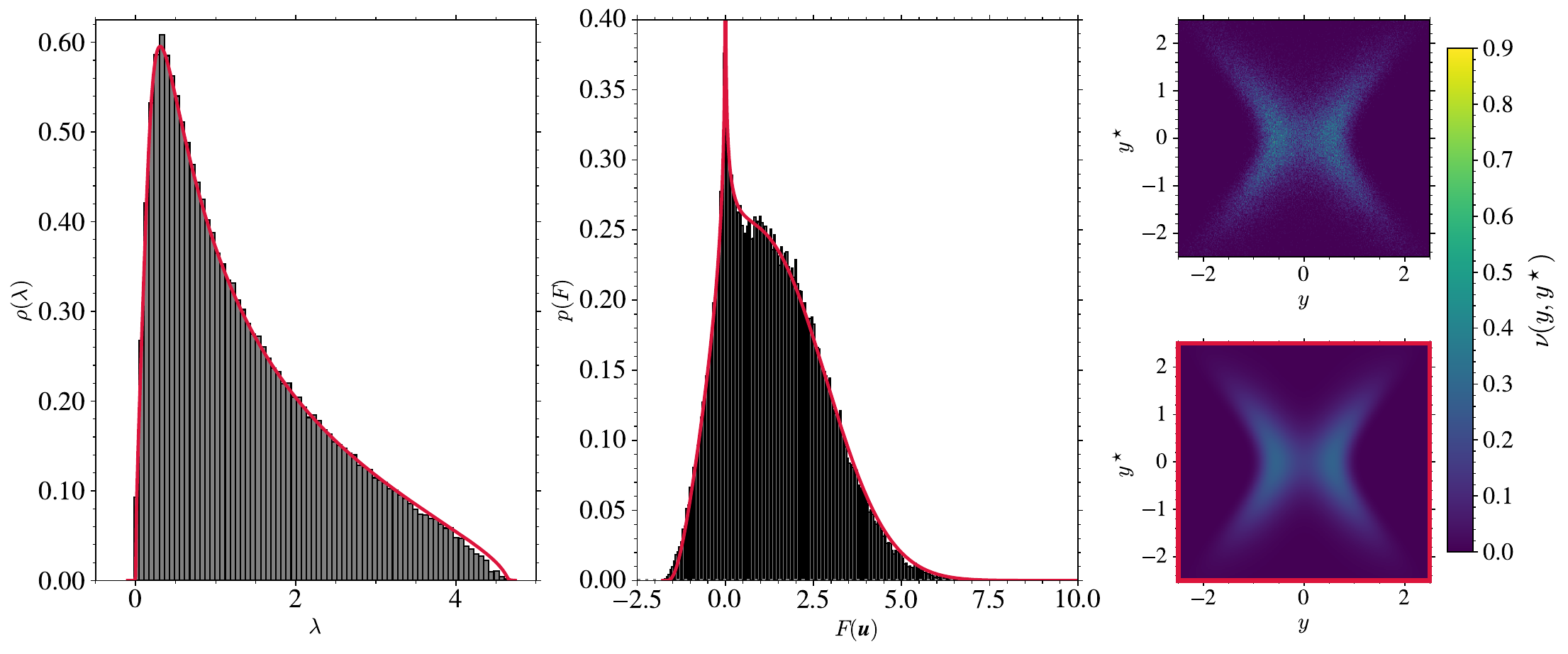}
   \caption{
        For $\alpha = 3.5$.
   }
  \end{subfigure}

  \medskip

  \begin{subfigure}{\linewidth}
    \centering
    \includegraphics[width=.9\linewidth]{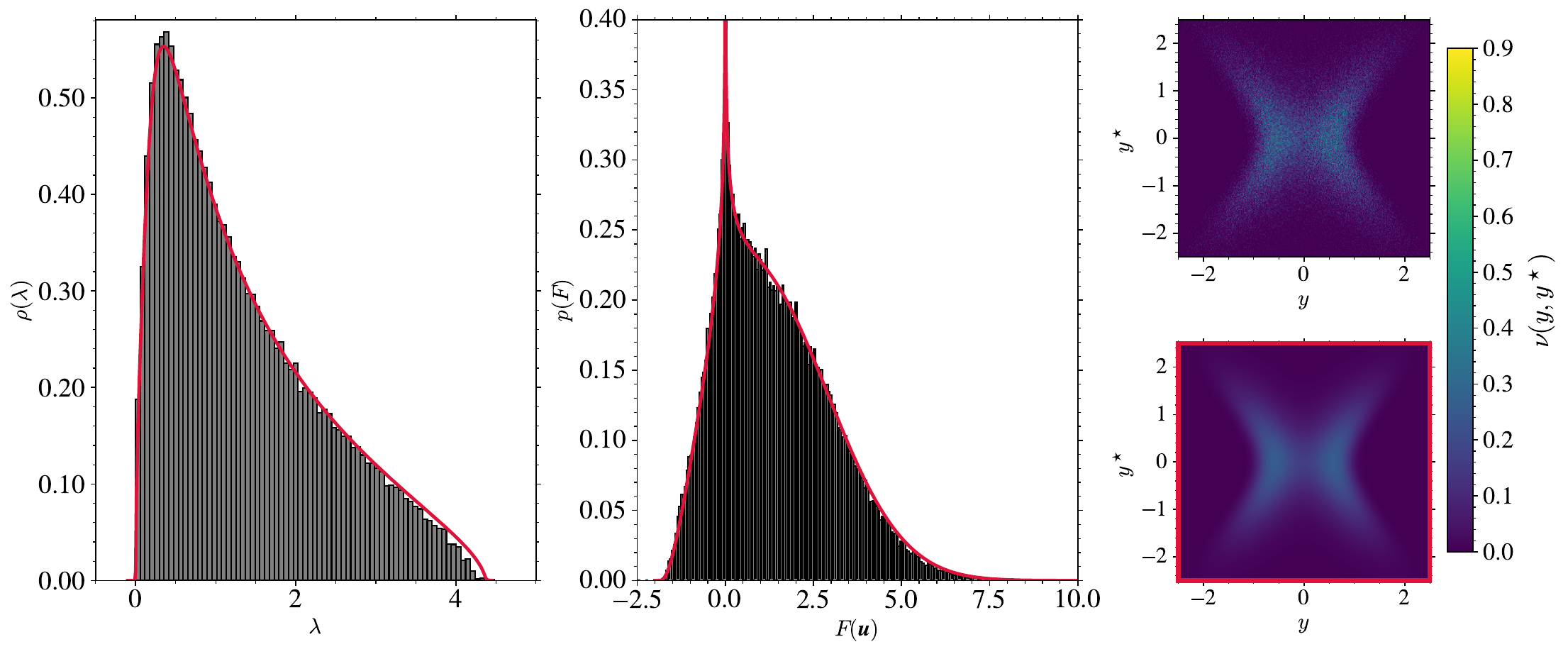}
   \caption{
        For $\alpha=4.5$.
   }
  \end{subfigure}
  
    \caption{
        Comparisons of the predicted Kac-Rice properties for the minima at $q=0.0$ and typical energy $e_\star$ and the empirical minima found by the gradient descent dynamics at $d=512$ and $a=1$. We compare: (left) the eigenvalue distribution of the Hessian $\rho(\lambda)$, (middle) the distribution of the Hessian weights $F(\bu)=\partial^2_1 \ell(y,y^\star)$, and (right) the joint label distributions $\nu(y, y^\star)$. In red are the Kac-Rice predictions described in Section~\ref{sec:kac_rice}.
        \label{fig:comparison_dynamics_H_nu_PF_a1}
    }
\end{figure}

While the main text focuses on comparing results from gradient descent dynamics at fixed loss normalization $a=0.01$ in eq.~\eqref{eq:def_ell_a}, the present appendix reproduces the comparison from Section~\ref{subsec:numerical_comparison} of the minima properties with the Kac-Rice prediction for $a=1$. In Fig.~\ref{fig:energies_a1.0_KR_exp} we show the comparison of the predicted and observed energies for the minima in the landscape, showing again a very good agreement with the typical energy predicted by the annealed Kac-Rice computation. In Fig.~\ref{fig:comparison_dynamics_H_nu_PF_a1}, we display the comparison at fixed $\alpha \in \{3.5, 4.5\}$ of the spectral properties of the minima, $F(\bu)$, and the signal-label joint probability distribution $\nu(y, y^\star)$. Again, the predictions at the typical energy fit remarkably well the shape of all the distributions in great details.

\textbf{A word about the BBP transition for $a=1$ --} For $q=0$, our theory predicts that $\alpha_\mathrm{BBP}^\mathrm{(typ)}=5.94$, whereas our $d=512$ simulations have their transition in the interval $\left[4.15, 5.20\right]$. However, we observe that the transition line for the success rate at $d=512$ for this value of $a$ is not sharp. Our definition for the critical $\alpha$ might not be suitable for this case. As an example, the more involved finite-size analysis from \cite{bonnaire2025role} reports a transition at $\alpha\approx 5.55$, much closer to our annealed prediction. We leave for a future version of this work the computationally-intensive task of evaluating the full phase diagram, similar to Fig.~\ref{fig:phase_diagram_exp_b},  for $a=1$.

